\documentclass[11pt]{article}

\usepackage[final]{acl}

\usepackage{times}
\usepackage{latexsym}

\usepackage[T1]{fontenc}
\usepackage[utf8]{inputenc}

\usepackage{microtype}

\usepackage{inconsolata}

\usepackage{graphicx}
\usepackage{multirow}
\usepackage{tabularx}
\usepackage{amsmath}
\usepackage{graphicx}
\usepackage{subfig}
\usepackage{soul}
\usepackage{enumitem}
\usepackage{multirow} % 支持表格中的 \multirow
\usepackage{booktabs} % 更专业的表格格式，如 \hline, \toprule 等
\usepackage{caption} % 处理表格和图片的标题
\usepackage{subfig}
\usepackage{xcolor}
\usepackage{amssymb}
\title{Negative-Aware Diffusion Process\\ for Temporal Knowledge Graph Extrapolation}

\author{Yanglei Gan, Peng He, Yuxiang Cai, Run Lin, Guanyu Zhou, Qiao Liu\thanks{Corresponding Author} \\
  \text{University of Electronic Science and Technology of China} \\
  \text{\{yangleigan, hepenglk, yuxiangcai, runlin, 202522080824\}@std.uestc.edu.cn,} \\
  \text{qliu@uestc.edu.cn} \\
  }

\begin{document}
\maketitle
\begin{abstract}
Temporal Knowledge Graph (TKG) reasoning seeks to predict future missing facts from historical evidence. While diffusion models (DM) have recently gained attention for their ability to capture complex predictive distributions, two gaps remain: (i) the generative path is conditioned only on positive evidence, overlooking informative negative context, and (ii) training objectives are dominated by cross-entropy ranking, which improves candidate ordering but provides little supervision over the calibration of the denoised embedding. To bridge this gap, we introduce \textbf{N}egative-\textbf{A}ware \textbf{D}iffusion model for TKG \textbf{Ex}trapolation (\textbf{NADEx}). Specifically, NADEx encodes subject-centric histories of entities, relations and temporal intervals into sequential embeddings. NADEx perturbs the query object in the forward process and reconstructs it in reverse with a Transformer denoiser conditioned on the temporal-relational context. We further derive a cosine-alignment regularizer derived from batch-wise negative prototypes, which tightens the decision boundary against implausible candidates. Comprehensive experiments on four public TKG benchmarks demonstrate that NADEx delivers state-of-the-art performance\footnote{The source code is online at: \url{https://github.com/AONE-NLP/TKG-NADEx}.}.
\end{abstract}

\section{Introduction}
Temporal Knowledge Graphs (TKGs) serve as dynamic structures representing entities and their evolving relationships through time \cite{ji2021survey,liang2024survey}, expressed via quadruples of the form $(s, r, o, t)$, such as (\textit{Donald Trump}, \textit{meet}, \textit{Joseph Biden}, 2024/12/21). Reasoning over TKGs predicts unknown links from historical observations and comprises two settings: \textbf{interpolation} \cite{cai2023temporal,luo2024chain}, which infers missing facts within the observed timeline, and \textbf{extrapolation}, which forecasts events beyond it \cite{yang2015embedding,zhu2021learning,Li21Temporal}. We focus on extrapolation as it provides anticipatory insight for future-oriented decision-making.

Accurate forecasting of future events necessitates a deep and nuanced comprehension of historical event sequences and relational dynamics. Prior research efforts \cite{jin2020recurrent,Li21Temporal,trivedi2017know} commonly adopt deterministic embedding strategies by integrating neighboring structural contexts and temporal dependencies. These representations are subsequently utilized within scoring functions such as TransE \cite{bordes2013translating}, DistMult \cite{yang2015embedding}, to assess the plausibility of prospective facts. 

Recent progress in TKG reasoning has largely followed a \textbf{learn‑to‑classify} paradigm, seeking ever‑richer representations while maintaining deterministic prediction targets. One prominent line of research augments graph neural networks (GNNs) with temporal gates that selectively transmit historical information, enabling fine-grained updates to entity and relation embeddings \cite{Li21Temporal,li2022tirgn}. A complementary stream adopts contrastive objectives, sharpening decision boundaries by contrasting local versus global neighborhoods \cite{chen2024local, pang2025improving} or historical versus non-historical contexts \cite{xu2023temporal}. Rule-based techniques further enhance interpretability by injecting symbolic priors into embedding models \cite{chen2025enhancing,chen2025cogntke}, while reinforcement-learning approaches guide multi-hop exploration with temporally calibrated rewards \cite{sun2021timetraveler,zheng2023dream}.

\begin{figure}[t]
        \centering
	\includegraphics[scale=0.48]{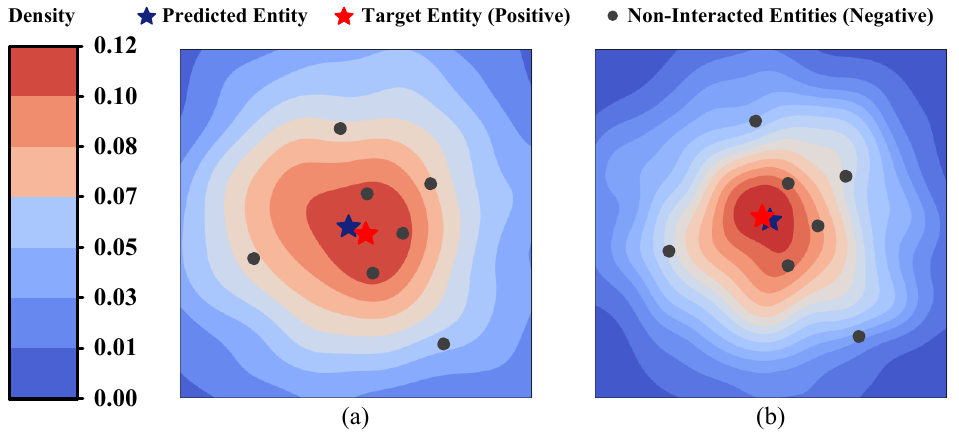}
        \caption{Illustration of entity embedding distributions learned by Diffusion-based TKG reasoning framework. (a) Neglecting the negative item distribution leads to predicted entities potentially drifting toward non-interacted negative entities. (b) Incorporating the negative sampling sharpens the predictive distribution, positioning the predicted entity centroid closer to the true target.}
    \label{fig1}
\end{figure}

Although these techniques secure strong empirical results, their deterministic nature limits their ability to model the inherent uncertainty of future events. Motivated by the generative prowess of diffusion models (DMs), recent work has reframed TKG reasoning as a \textbf{learn-to-generate} paradigm, exploiting DMs to sample plausible target object \cite{cai-etal-2024-predicting,chen2025llm,cao2025dpcl}. Despite their empirical success, existing methods exhibit two key limitations: 

\begin{itemize}[left=0em] 
    \item \textbf{Neglect of Negative Context in Query Conditioning. } Existing DM-based methods generate entity embeddings by conditioning solely on observed positive facts, without accounting for negative examples that are essential for discrimination \cite{xu2023temporal,chen2024local,cao2025dpcl}. As illustrated in Figure \ref{fig1}(b), explicitly incorporating negative samples into the generative diffusion process introduces counterfactual evidence that regularizes the trajectory and reduces false-positive bias, yielding a lower-variance predictive distribution that is more tightly centered on true positives.
    \item \textbf{Reliance on Generic Loss Formulations. } Most existing work optimizes diffusion dynamics with generic reconstruction losses such as vanilla cross-entropy ranking \cite{cai-etal-2024-predicting}, that emphasize relative ordering of candidate entities rather than enforcing explicit margins separating plausible from implausible outcomes \cite{chen2025llm}. Therefore, the generative potential of diffusion remains under-exploited.
\end{itemize}

In this paper, we propose a \textbf{N}egative-\textbf{A}ware \textbf{D}iffusion model for temporal knowledge graph \textbf{Ex}trapolation (NADEx). Specifically, we reformulate temporal knowledge graph (TKG) reasoning as a sequence prediction task by encoding historical objects, event types, and temporal intervals into sequential embeddings. To capture uncertainty in future events, Gaussian noise is injected into target entity embeddings, and a Transformer-based denoising network conditioned on historical contexts and temporal dynamics reconstructs these embeddings. Moreover, we develop a cosine-alignment regularizer for diffusion-based TKG reasoning that leverages batch-wise negative prototypes, computed by averaging the embeddings of the other targets in the mini-batch. Used alongside cross-entropy reconstruction loss, this direction-aware constraint sharpens separation between plausible and implausible candidate entities. Our contributions are three-fold:

\begin{itemize}[leftmargin=*]
    \item We propose NADEx, a negative‑aware diffusion model for temporal knowledge graph extrapolation, explicitly capturing the dynamic and stochastic nature of future events through a Gaussian perturbed sequence denoising process. 
    \item NADEx optimizes a cross-entropy reconstruction objective augmented with a cosine-alignment regularizer, sharpening separation between plausible and implausible candidate entities thus yielding a more discriminative predictive distribution.
    \item Extensive experiments conducted on four real world datasets demonstrate that NADEx yields superior performance, while remaining robust across diverse evaluation scenarios.
\end{itemize}

\section{Related Works}

\subsection{Discriminative TKG Reasoning}

Temporal Knowledge Graph (TKG) reasoning seeks to infer future facts by leveraging the chronological trajectory of previously observed triples. Early work such as Know-Evolve \cite{bordes2013translating, chen2024thcn} treats the TKG as a set of mutually-exciting Hawkes processes, yielding continuous-time intensity functions for entity–relation interactions. THCN \cite{chen2024thcn} couples a Hawkes process with a temporal causal convolutional network, enabling causal convolutions to inherit the self-exciting inductive bias of point processes. However, these methods propagate information only to adjacent timestamps, leaving long-range dependencies under-exploited. Subsequent studies enhance static KG encoders with temporal signals. RE-NET \cite{jin2020recurrent}, RE-GCN \cite{Li21Temporal}, and CyGNet \cite{zhu2021learning} inject recurrent gating, temporal self-attention, and cycle-aware constraints, respectively, to capture evolving patterns, while structural extensions, CEN \cite{li2022complex}, xERTE \cite{han2021explainable} and HisMatch \cite{li2022hismatch} exploit hierarchical alignment for richer representations. Yet their reliance on single-scale memories still limits sensitivity to distant context.

Considering the long-term dependencies among entities and relations, Zhang et al. \cite{zhang2023learning} decompose each entity–relation trajectory into long- and short-term streams. Building on that idea, CENET \cite{xu2023temporal} formulates temporal reasoning as a history-aware contrastive task: for every query, positive events are contrasted against hard negatives drawn from both historical and non-historical contexts. LogCL \cite{chen2024local} further refines this idea with a local–global paradigm that first selects query-relevant snapshot windows via entity-centric attention and then applies four bespoke contrastive patterns.  

Parallel to these efforts, symbolic and structural approaches pursue interpretability and inductive generalization. Rule-based systems such as Tlogic \cite{liu2022tlogic} mine time-stamped first-order clauses via temporal random walks, offering interpretable but coverage-limited priors. DaeMon \cite{dong2023adaptive} aggregate relation-specific temporal paths between query entities, replacing brittle entity embeddings with structural evidence that generalizes to unseen nodes. CognTKE \cite{chen2025cogntke} further unifies shallow reasoning and local deep path reasoning over a cognitive temporal relation digraph, achieving notable gains. 

\subsection{Generative TKG Reasoning}

Most recently, generative frameworks have introduced principled uncertainty modelling. DiffuTKG \cite{cai-etal-2024-predicting} recasts forecasting as conditional sequence denoising under a forward-and-reverse Gaussian diffusion process, and DPCL-Diff \cite{cao2025dpcl} couples graph-node diffusion with dual-domain periodic contrastive learning to disentangle recurrent cycles from genuinely novel events. Complementing these methods, Luo et al. \cite{luo2024chain} leverage large language models (LLMs) to generate multi-step event chains that capture higher-order dependencies, while LLM-DR \cite{chen2025llm} harnesses a classifier-free guided diffusion model conditioned on temporal KG contexts and refines generated candidate rules with LLM-based constraints. Despite these advances, existing methods often overlook negative contexts and rely on generic loss formulations, resulting in blurred predictive distributions. Such limitations underscore the need for a generative framework that can sharply discriminate future events.

\section{Preliminary}

\textbf{Definition 1. Temporal Knowledge Graph. } Let $\mathcal{E}$, $\mathcal{R}$, and $\mathcal{T}$ denote finite sets of entities, relation types, and timestamps, respectively. A temporal knowledge graph $\mathcal{G}$ is a collection of time-stamped quadruples:
\begin{equation}
\footnotesize
\mathcal{G}
   \;=\;
   \Bigl\{\, (s,r,o,t) \;\Big|\;
  s,o \in \mathcal{E},\;
  r \in \mathcal{R},\;
  t \in \mathcal{T}
   \Bigr\},
\label{eq:tkg_definition}
\end{equation}
where each tuple encodes the fact that relation $r$ holds from subject $s$ to object $o$ at time $t$. More specifically, the TKG can be viewed as an ordered sequence of static snapshots:
\begin{equation}
\footnotesize
\mathcal{G}
   \;=\;
   \bigl\{
      \mathcal{G}_{1},
      \mathcal{G}_{2},
      \ldots,
      \mathcal{G}_{|\mathcal{T}|}
   \bigr\},
\qquad
\mathcal{G}_{t}\;\subseteq\;
   \mathcal{E}\times\mathcal{R}\times\mathcal{T},
\label{eq:snapshot_sequence}
\end{equation}
where $\mathcal{G}_{t}$ aggregates all triples that are valid at timestamp $t$. Following the standard bidirectional–relation convention \cite{kazemi2018simple}, we augment every quadruple $(s, r, o, t)$ with its inverse $(o, r^{-1}, s, t)$, where $r^{-1}$ is a distinct relation denoting the reverse semantics of $r$. 

\noindent\textbf{Definition 2. Temporal Knowledge Graph Reasoning. } Let $q = (s, r, ?, t)$ be a query quadruple whose object entity is missing at timestamp $t$. Given the sliding history window of length $L$, $\mathcal{G}_{t-L-1:t-1} = \{\mathcal{G}_{t-L},\mathcal{G}_{t-L+1},\dots,\mathcal{G}_{t-1}\}$. The TKG reasoning task is to learn a scoring function:

\begin{equation}
\footnotesize
\operatorname{score}_t(o)=f(s, r, o, \mathcal{G}),\\
\hat{o}=\underset{o \in \mathcal{E}}{\arg \max } \operatorname{score}_t(o) .
\end{equation}

Each candidate object \(o\in\mathcal{E}\) is assigned a plausibility score and the highest‐scoring entity completes the quadruple.

\section{Method}

% In this section, we detail our framework, illustrated in Figure \ref{fig2}, which comprises three main modules: (1) Temporal Representation Learning encodes timestamps as contextual graphs, capturing semantic interactions among co-occurring entities; (2) Forward Diffusion progressively injects Gaussian noise into object embeddings, modeling generative uncertainty; (3) Reverse Denoising iteratively recovers original object embeddings from noisy inputs, guided by temporal context and relational information. Finally, we formalize our training objective and outline the inference process.

\begin{figure*}[t]
        \centering
	\includegraphics[scale=0.95]{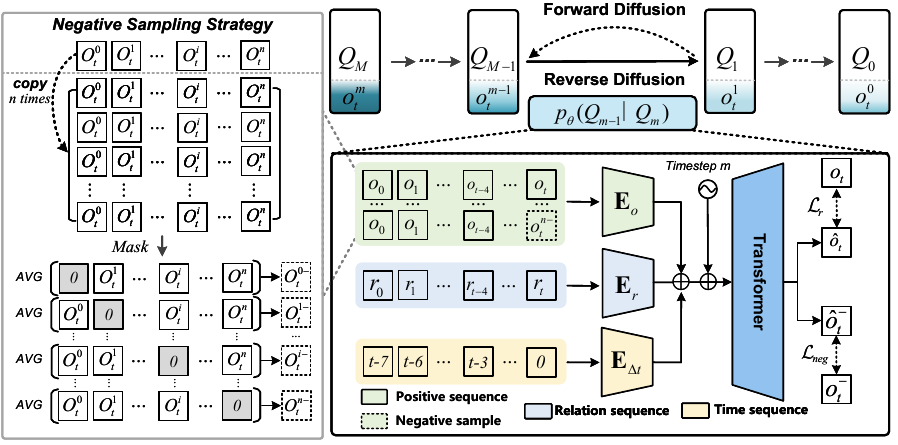}
        \caption{Overview of NADEx. In Forward Diffusion, Gaussian noise is applied to the future object $o$. During the reverse denoising process, NADEx incorporates both positive and negative ($o_{t}^{n-}$) samples to disentangle plausible from implausible event predictions. For negative samples, we first stack all target embeddings in the mini-batch into a single row, copy this row $n$ times to form an $n$ $\times$ $n$ matrix, set the diagonal to 0 to exclude each target itself, and then take the row-wise mean of the remaining entries to obtain one compact negative prototype per target.}
    \label{fig2*}
\end{figure*}

\subsection{Temporal Representation Learning}
Given a query $q=(s,r,?,t)$, we follow \citet{cai-etal-2024-predicting} and cast TKG extrapolation as sequence prediction over the subject's history. Let the subject-centric history up to time $t-1$ be $Q_{0:t-1} = \{(s,r_0,o_0,t_0),\dots,(s,r_i,o_i,t_i),\dots,$ $(s,r_{t-1},o_{t-1},t_{t-1})$\}, where $L$ is the window length. We form three aligned sequences $O_{0:t-1}$, $R_{0:t-1}$, $T_{0:t-1}$, and employ three trainable matrices to get their corresponding embeddings,

\begin{equation}
\footnotesize
\begin{aligned}
\mathbf{O}^{+} = \mathbf{E}_o\left[\mathbf{o}_0; \, \mathbf{o}_1; \, \ldots; \, \mathbf{o}_{t-1}, \mathbf{o}_{t}\right]; \\
\mathbf{R} = \mathbf{E}_r\left[\mathbf{r}_0; \, \mathbf{r}_1; \, \ldots; \, \mathbf{r}_{t-1}, \mathbf{r}_{t}\right]; \\
\mathbf{T} = \mathbf{E}_{\Delta t}\left[\mathbf{t}_0; \, \mathbf{t}_1; \, \ldots; \mathbf{t}_{t-1}, \mathbf{t}_t\right].
\end{aligned}
\end{equation}

where $\mathbf{E}_o \in \mathbb{R}^{\mathcal{E}\times h}$, $\mathbf{E}_r \in \mathbb{R}^{\mathcal{R}\times h}$, $\mathbf{E}_{\Delta t} \in \mathbb{R}^{t \times h}$. $\mathcal{E}$, $\mathcal{R}$, and $h$ denotes the number of entities, event types and the embedding dimension, respectively. $\bold{O}_{0:t-1}$ denotes the embeddings of the historical object entities, denoted as $\mathcal{H}=\mathbf{O}_{0:t-1}$ for brevity, while $\bold{o}_t \in \mathbb{R}^{1 \times h}$ refers to the representation of the target entity to be predicted.

% \begin{equation}
% \footnotesize
% \mathbf{o}_i &= \mathbf{E}_o[o_i], \quad \mathbf{r}_i &= \mathbf{E}_r[r_i], \quad \mathbf{t}_i &= \mathbf{E}_{\Delta}[t_i].
% \end{equation}
% $\mathbf{E}_o$ captures the historical object embeddings, $\mathbf{E}_r$ encodes their corresponding event types, and $\mathbf{E}_\Delta$ models the relative temporal distances to the query time $t$. These vectors are then stacked to form the sequence representations:
% \begin{equation}
% \footnotesize
% \begin{aligned}
% \mathbf{O} = \left[\mathbf{o}_0; \, \mathbf{o}_1; \, \ldots; \, \mathbf{o}_{t-1}, \mathbf{o}_{t}\right]; \\
% \mathbf{R} = \left[\mathbf{r}_0; \, \mathbf{r}_1; \, \ldots; \, \mathbf{r}_{t-1}, \mathbf{r}_{t}\right]; \\
% \mathbf{T} = \left[\mathbf{t}_0; \, \mathbf{t}_1; \, \ldots; \mathbf{t}_{t-1}, \mathbf{t}_t\right].
% \end{aligned}
% \end{equation}
% where $\bold{O}_{0:t-1}$ denotes the embeddings of the historical object entities, denoted as $\mathcal{H}=\mathbf{O}_{0:t-1}$ for brevity, while $\bold{o}_t \in \mathbb{R}^{1 \times h}$ refers to the representation of the target entity to be predicted.

% \begin{figure}[t]
% \centering
% 	\includegraphics[scale=0.88]{模型图gan2.png}
% \caption{Overview of the negative sample generation process. For each mini-batch, we duplicate each target sequence $o_{{0:n} ,t}$ $N$ times to form a candidate pool, where $N$ is the batch size. Within each copy, we mask the object in grey blocks and compute the average embedding of the remaining items to produce a synthetic “negative” target object.}
%     \label{fig3}
% \end{figure}
\subsection{Negative Sampling Strategy}
To exploit the discriminative signal from negative evidence, we design a batch-wise negative sampling strategy. For a mini-batch of size $n$, we first collect the target entities into a sequence $O_{t} = \{o_{t}^{0},o_{t}^{1},\dots,o_{t}^{n}\}\in \mathbb{R}^{n \times 1}$, and then construct an $n \times n$ mask by permuting the diagonal entries, resulting in each row containing the other $N-1$ objects from the batch as negatives. \textbf{Averaging} these negative candidates along each row generates a compact negative sequence, defined as:

\begin{equation}
\tiny
\setlength{\arraycolsep}{4pt}
\renewcommand\arraystretch{1.3}
\begin{pmatrix}
o_{t}^{0} & o_{t}^{1} &\cdots& o_{t}^{n}      \\
o_{t}^{0} & o_{t}^{1}      & \cdots & o_{t}^{n}      \\
\vdots & \vdots & \ddots & \vdots \\
o_{t}^{0} & o_{t}^{1}      & \cdots & o_{t}^{n}
\end{pmatrix}
\Rightarrow
AVG.
\left\{
\begin{gathered}
\begin{pmatrix}
0 & o_{t}^{1}      & \cdots & o_{t}^{n} 
\end{pmatrix}\\
\begin{pmatrix}
o_{t}^{0} & 0      & \cdots & o_{t}^{n} 
\end{pmatrix}\\
\begin{pmatrix}
\makebox[1.35em]{$\vdots$} & \vdots & \ddots & \makebox[1.35em]{$\vdots$}
\end{pmatrix}\\
\begin{pmatrix}
o_{t^{0}} & o_{t}^{1}      & \cdots & 0
\end{pmatrix}
\end{gathered}
\right\}
=
\begin{pmatrix}
o_{t}^{0-} \\
o_{t}^{1-}\\
\vdots \\
o_{t}^{n-}
\end{pmatrix}.
\end{equation}
% Where each aggregated negative object $o_{it}^{neg}$ is computed by averaging the negative candidates in the corresponding row:
% \begin{equation}
% \footnotesize
% o_{it}^{neg} = AVG(o_{i0}^{neg},\dots,o_{i(i-1)}^{neg},o_{i(i+1)}^{neg},\dots,o_{it}^{neg}).
% \end{equation}
The treatment of negative samples $o_{t}^{-}$ is equivalent to $o_{t}$, which is obtained through the entity embedding matrix $\mathbf{E_o}$:
% \begin{equation}
% \mathbf{O}_t^{neg} = \mathbf{E_o}[o_{0t}^{neg};\cdots; o_{it}^{neg}; \cdots ;o_{nt}^{neg}].    
% \end{equation} 
\begin{equation}
\footnotesize
\mathbf{o}_t^{-} = \mathbf{E_o}[o_{t}^{-}]. 
\end{equation} 
Finally, our negative sequence can be expressed as:
\begin{equation}
\footnotesize
\mathbf{O}^{-} = \left[\mathbf{o}_0; \, \mathbf{o}_1; \, \ldots; \, \mathbf{o}_{t-1}, \mathbf{o}_{t}^{-}\right].
\end{equation}

It is important to note that, due to the event-stream formulation of temporal knowledge graph (TKG) extrapolation and the way events are organized in our benchmarks, we treat all events sharing the same timestamp as a single training unit. Concretely, if $N$ events occur at a given timestep, we construct one full batch that contains all $N$ corresponding target entities. This design follows directly from the standard TKG reasoning objective: at timestamp $t+1$, the model is required to predict the missing object entities for all queries occurring at that time. Consequently, $N$ is not a tunable batch-size hyperparameter but is uniquely determined by the number of real-world events recorded at each timestamp, which can vary substantially across time.

\subsection{Forward Diffusion Process}

We model predictive uncertainty by diffusing only the candidate object embedding $\bold{o}_t$ and sampled negatives $\bold{o}_t^{-}$. The forward process progressively add Gaussian noise over $M$ steps:

\begin{equation}
\footnotesize
\begin{aligned}  
o_m^+=\sqrt{\overline{\alpha}_m}o_t+\sqrt{1-\overline{\alpha}_m}\boldsymbol{\epsilon},\\ 
o_m^-=\sqrt{\overline{\alpha}_m}o_t^{-}+\sqrt{1-\overline{\alpha}_m}\boldsymbol{\epsilon},
\end{aligned}
\end{equation}
where $\boldsymbol{\epsilon} \in \mathcal{N}(0, \mathbf{I})$ introduces Gaussian noise, and $\bar{\alpha}_{m} \in (0,1)$ controls the signal-to-noise ratio at step $m$, simulating the predictive uncertainty.

To regulate the accumulation of noise over time, we employ a linear schedule that defines 1-$\bar{\alpha}_{m}$ as:

\begin{equation}
\label{eq:linear_schedule}
\footnotesize
1 - \bar{\alpha}_m = \delta \cdot \left(\alpha_{\min} + \frac{m - 1}{M - 1} (\alpha_{\max} - \alpha_{\min})\right).
\end{equation}
where $\delta \in [0,1]$ is a global scaling factor that moderates the overall diffusion strength, and $\alpha_{min}$ and $\alpha_{max}$ are hyperparameters that determine the lower and upper bound of noise levels. 

% This schedule ensures a smooth and controllable corruption process, allowing the model to gradually learn how to reconstruct the clean embedding from increasingly noisy inputs during the reverse denoising phase.

% Meanwhile, our query object can be represented as $\mathbf{O}^{+}=[\mathcal{H},o_m^+]$, $\mathbf{O}^{-}=[\mathcal{H},o_m^-]$, where $\mathbf{O}^{+}, \mathbf{O}^{-}\in \mathbb{R}^{N \times |\mathcal{O}| \times h}$. Further, we fuse the information of $\mathbf{O}^{+}$, $\mathbf{O}^{-}$, $\mathbf{R}$ and $\mathbf{T}$, denoted as:

\subsection{Reverse Denoising Process}
In reverse process, we aim to recover the target sequence of object entities representation $o_t$ or $o_t^{-}$  iteratively from a pure Gaussian noise $o_m^{+}$ or $o_m^{-}$. At each step, the model is conditioned on the temporal $\mathbf{T}$ and relational $\mathbf{R}$ context: 

\begin{equation}
\footnotesize
p_\theta(\hat{\mathbf{o}}_{m-1}|*)=\mathcal{N}\left(\hat{\mathbf{o}}_{m-1};\mu_\theta(*),\Sigma_\theta(*)\right).
\end{equation}
where we use a Transformer to model $\mu_\theta(*)$ and $\Sigma_\theta(*)$ during reverse process. For brevity, $*$ denotes [$\mathbf{\hat{o}}_{m}, \mathbf{r}, \mathbf{t}, m$], which is the denoising conditions. $\hat{o}_{m}$ is set to $\mathcal{H}$ for the first step:

\begin{equation}
\footnotesize
\begin{aligned}
\hat{\mathbf{o}}_{0}^{+}=f_\theta\left(\mathbf{Z}_x\right)=\text {Transformer}\left(\left[\mathbf{o}_{i}^{+}+\operatorname{Emb}(i)\right]\right),\\
\hat{\mathbf{o}}_{0}^{-}=f_\theta\left(\mathbf{Z}_x\right)=\text {Transformer}\left(\left[\mathbf{o}_{i}^{-}+\operatorname{Emb}(i)\right]\right).
\end{aligned}
\end{equation}
where $i\in[0,m]$, $\text{Emb}(\cdot)$ denotes denoising step embeddings to manage the hidden representations at different noise levels.

% \begin{equation}
% \footnotesize
% \begin{aligned}
% \mathbf{Q}^{+} = \mathbf{LN}(\mathbf{Dropout}(\mathbf{O}^{+}+\mathbf{R}+\mathbf{T})),\\
% \mathbf{Q}^{-} = \mathbf{LN}(\mathbf{Dropout}(\mathbf{O}^{-}+\mathbf{R}+\mathbf{T})).
% \end{aligned}
% \end{equation}
% Where $\mathbf{LN}$ denotes LayerNorm, $\mathbf{Q}^{+}$ and $\mathbf{Q}^{-}$ denote positive and negative samples, respectively. 

% The denoising process are as follows:

% \begin{equation}
% \footnotesize
% p_\theta(\mathbf{Q}_{0:M})=p(\mathbf{Q}_M)\prod_{m=1}^Mp_\theta(\mathbf{Q}_{m-1}|\mathbf{Q}_m),
% \end{equation}
% we use a learnable approximator $f_{\theta}(*)$ to model the reverse process at each step:

% where we denote $*$ to represent [$\mathbf{\hat{Q}}_{m}, m$] for brevity. NADEx utilizes Transformer as the backbone to recover $\mathbf{\hat{Q}}_{0}$ during the reverse phase:

% \begin{equation}\begin{aligned}
% \footnotesize
% \mathbf{\hat{Q}}_0=f_\theta(\mathbf{Z}_x)=\text{Transformer }([\mathbf{z}_0,\mathbf{z}_1,..,\mathbf{z}_m]),\\
%  \mathbf{z}_i=\mathbf{Q}_i+\text{Emb}(i),
% \end{aligned}\end{equation}

\subsection{Training Objective}

\textbf{Reconstruction Loss.}
We introduce a distance quantification measurement based on the dot product between embeddings:

\begin{equation}
\footnotesize
\mathbf{Y}=\text{Softmax}(\mathbf{\hat{o}}_0^+\cdot(\mathbf{E})^T),
\end{equation}
where $\mathbf{\hat{o}}_0^+\in \mathbb{R}^{1 \times h}$ denotes the representation of the target object from the reverse process. $(\cdot)^T$ is the matrix transposition operation. Moreover, we utilize a reconstruction loss function $\mathcal{L}_{\mathrm{r}}$ as follows:
\begin{equation}
\footnotesize
\mathcal{L}_{\mathrm{r}}=-\sum_{i\in\{1,2,...,d\}}y_i\log(\mathbf{
Y}_i),
\label{l1}
\end{equation}
where $y_i$ denotes the one-hot encoding of the $i$-th gold entity, and $\mathbf{Y}_i$ is the predicted probability.

\noindent\textbf{Negative Sampling Loss.} Exclusively depending on the generation of optimization objectives may result in model overfitting to historical frequent events. Though this method can alleviate the impact of sparse and noisy interactions inherent in the system, it might fail to fully harness the generative and generalization capacities of diffusion models, thereby leading to inaccurate assessments of both unseen and observed facts. To gain a better understanding and to specifically redesign for modeling object distributions in order to achieve diffusion optimization goals. Specifically, our negative sample cosine loss ($\mathcal{L}_{\mathrm{neg}}$) is represented as:

\begin{equation}
\footnotesize
\mathcal{L}_{\mathrm{neg}}=\frac{1}{N}\sum_{i=1}^N\left(\frac{\mathbf{o}_i^-}{\|\mathbf{o}_i^-\|_2}\cdot\frac{\hat{\mathbf{o}}_i^-}{\|\hat{\mathbf{o}}_i^-\|_2}-1\right)^2.
\label{l2}
\end{equation}

% Subsequently, through Eq. \eqref{l1} and \eqref{l2}, the loss deviation is calculated as:
% \begin{equation}
% \footnotesize
% \mathcal{L}_{r\_neg}=\mathcal{L}_{\mathrm{r}}-\mathcal{L}_{\mathrm{neg}}.
% \end{equation}

% Finally, the combination of the negative logarithmic $\text{Sigmoid}$ term and the weighted diffusion loss is expressed as:
The training objective can be formulated as:
% \begin{equation}
% \mathcal{L}=-(1-\lambda)\log \sigma\left(\frac{1}{1+e^{\gamma \mathcal{L}_{r\_neg} -\epsilon}}\right)+\lambda \mathcal{L}_{\mathrm{r}},
% \end{equation}
\begin{equation}
\footnotesize
\mathcal{L}=-(1-\lambda)\log \sigma\left(-\gamma\cdot(\mathcal{L}_{r}-\mathcal{L}_{neg})+\epsilon\right)+\lambda \mathcal{L}_{\mathrm{r}}.
\end{equation}
where $\sigma$ is the $\text{Sigmoid}$ function, $\epsilon = 1e-8$  to prevent numerical instability, and $\lambda$ and $\gamma$ are equilibrium weight hyperparameters.
% \mathcal{L}_{r}+

\begin{table*}[t]
\caption{Overall performance comparison (\%) on four datasets in terms of MRR and Hit@1/3/10 with time-aware metrics. The best results are highlighted in \textbf{bold}, while the second-best results are \underline{underlined}. Statistical significance is determined via paired bootstrap $t$-test, with differences at $p<0.05$ deemed significant.}
\setlength{\tabcolsep}{1.5pt} % 减小列间距
\fontsize{8pt}{8pt}\selectfont % 设置特定字号和行距
\begin{tabular}{@{}ccccccccccccccccc@{}}
\toprule
\multirow{2}{*}{Models} & \multicolumn{4}{c}{ICEWS14}    & \multicolumn{4}{c}{ICEWS18}    & \multicolumn{4}{c}{ICEWS05-15} & \multicolumn{4}{c}{GDELT}      \\ \cmidrule(l){2-17} 
  & MRR   & Hit@1 & Hit@3 & Hit@10 & MRR   & Hit@1 & Hit@3 & Hit@10 & MRR   & Hit@1 & Hit@3 & Hit@10 & MRR   & Hit@1 & Hit@3 & Hit@10 \\ \midrule
DisMult \citeyearpar{yang2015embedding}  & 15.44 & 10.91 & 17.24 & 23.92  & 11.51 & 7.03  & 12.87 & 20.86  & 17.95 & 13.12 & 20.71 & 29.32  & 8.68  & 5.58  & 9.96  & 17.13  \\
ConvE \citeyearpar{dettmers2018convolutional}    & 35.09 & 25.23 & 39.38 & 54.68  & 24.51 & 16.23 & 29.25 & 44.51  & 33.81 & 24.78 & 39.00 & 54.95  & 16.55 & 11.02 & 18.88 & 31.60  \\
RotatE \citeyearpar{sun2018rotate}    & 21.31 & 10.26 & 24.35 & 44.75  & 12.78 & 4.01  & 14.89 & 31.91  & 24.71 & 13.22 & 29.04 & 48.16  & 13.45 & 6.95  & 14.09 & 25.99  \\ \midrule
TTransE \citeyearpar{leblay2018deriving}  & 13.72 & 2.98  & 17.70 & 35.74  & 8.31  & 1.92  & 8.56  & 21.89  & 15.57 & 4.80  & 19.24 & 38.29  & 5.50  & 0.47  & 4.94  & 15.25  \\
TA-DisMult \citeyearpar{garcia2018learning}  & 25.80 & 16.94 & 29.74 & 42.99  & 16.75 & 8.61  & 18.41 & 33.59  & 24.31 & 14.58 & 27.92 & 44.21  & 12.00 & 5.76  & 12.94 & 23.54  \\
DE-SimIE \citeyearpar{goel2020diachronic}  & 33.36 & 24.85 & 37.15 & 48.92  & 19.30 & 11.53 & 21.86 & 34.80  & 35.02 & 25.91 & 38.99 & 52.75  & 19.70 & 12.22 & 21.39 & 33.70  \\ \midrule
RE-NET \citeyearpar{jin2020recurrent}    & 36.93 & 26.83 & 39.51 & 54.78  & 28.81 & 19.05 & 32.44 & 47.51  & 43.32 & 33.43 & 47.77 & 63.06  & 19.62 & 12.42 & 21.00 & 34.01  \\
RE-GCN \citeyearpar{Li21Temporal}    & 40.39 & 30.66 & 44.96 & 59.21  & 30.58 & 21.01 & 34.34 & 48.75  & 48.03 & 37.33 & 53.85 & 68.27  & 19.64 & 12.42 & 20.90 & 33.69  \\
CyGNet \citeyearpar{zhu2021learning}    & 35.05 & 25.73 & 39.01 & 53.55  & 24.93 & 15.90 & 28.28 & 42.61  & 36.81 & 26.61 & 41.63 & 56.22  & 18.48 & 11.52 & 19.57 & 31.98  \\
TITer \citeyearpar{sun2021timetraveler}    & 41.73 & 32.74 & 46.46 & 58.44  & 29.98 & 22.05 & 33.46 & 44.83  & 47.69 & 37.95 & 52.92 & 65.81  & 15.46 & 10.98 & 15.61 & 24.31  \\
CEN \citeyearpar{li2022complex}       & 42.20 & 32.08 & 47.46 & 61.31  & 31.50 & 21.70 & 35.44 & 50.59  & 46.84 & 36.38 & 52.45 & 67.01  & 20.39 & 12.96 & 21.77 & 34.97  \\
TiRGN \citeyearpar{li2022tirgn}     & 44.04 & 33.83 & 48.95 & 63.84  & 33.66 & 23.19 & 37.99 & 54.22  & 50.04 & 39.25 & 56.13 & 70.71  & 21.67 & 13.63 & 23.27 & 37.60  \\
HisMatch \citeyearpar{li2022hismatch}  & 46.42 & 35.91 & 51.63 & \underline{66.84}  & 33.99 & 23.91 & 37.90 & 53.94  & \underline{52.85} & 42.01 & 59.05 & \textbf{73.28}  & 22.01 & 14.45 & 23.80 & 36.61  \\
% L2TKG     & 45.89 & 34.63 & --    & 68.47  & 31.63 & 21.17 & --    & 53.01  & 52.42 & 40.09 & --    & 75.86  & 20.16 & 12.49 & --    & 35.83  \\
RETIA \citeyearpar{liu2023retia}     & 42.76 & 32.28 & 47.77 & 62.75  & 32.43 & 22.23 & 36.48 & 52.94  & 47.26 & 36.64 & 52.90 & 67.76  & 20.12 & 12.76 & 21.45 & 34.49  \\
CENET \citeyearpar{xu2023temporal}     & 39.02 & 29.62 & 43.23 & 57.49  & 27.85 & 18.15 & 31.63 & 46.98  & 41.95 & 32.17 & 46.93 & 60.43  & 20.23 & 12.69 & 21.70 & 34.92  \\
CRAFT \citeyearpar{zhang2024modeling}     & 45.71 & 35.05 & 51.83 & 65.21  & 34.21 & 23.96 & 38.53 & 54.11  & 50.14 & 39.56 & 56.18 & 70.09  & \textbf{23.78} & \underline{15.38} & \underline{26.23} & \underline{40.15}  \\
THCN \citeyearpar{chen2024thcn}      & 45.39 & \underline{36.58} & 50.84 & 66.07  & 35.63 & 24.90 & 39.26 & 56.76  & 51.94 & 40.32 & 57.79 & 72.18  & 23.46 & 15.18 & 25.21 & 39.03  \\
DiffuTKG \citeyearpar{cai-etal-2024-predicting} & \underline{47.58}      &   36.38    &    \underline{53.41}   &     66.01   &      \underline{35.65} &     \underline{27.19}  &   \underline{44.04}    &     \underline{59.55}   &     48.97  &   39.80    &  56.92     &    69.84    &   21.87    &  14.43     &   23.68    &  36.05      \\
% LogCL \citeyearpar{chen2024local}     & 48.87 & 37.76 & 54.71 & 70.26  & 35.67 & 24.53 & 40.32 & 57.74  & 57.04 & 46.07 & 63.72 & 77.87  & \textbf{23.75} & 14.64 & 25.60 & 42.33  \\
LogiQ \citeyearpar{chen2025enhancing}     & 44.71 & 35.72 & 51.03 & 64.21  & 34.94 & 24.76 & 39.57 & 56.32  & 51.04 & 40.71 & 57.55 & 71.00  & -- & -- & -- & --  \\
CognTKE \citeyearpar{chen2025cogntke}     & 46.06 & 36.49 & 51.11 & 64.49  & 35.24 & 25.21 & 39.93 & 54.71  & \textbf{53.13} & \underline{42.62} & \underline{59.42} & \underline{72.70}  & -- & -- & -- & --  \\\midrule
% Llama-2-7B \citeyearpar{luo2024chain}     & -- & 34.90 & 47.00 & 59.10  & -- & 22.30 & 36.30 & 52.20  & -- & 38.60 & 54.10 & 69.90  & -- & -- & -- & --  \\ 
% Vicuna-7B \citeyearpar{luo2024chain}     & -- & 32.80 & 45.70 & 65.60  & -- & 20.90 & 34.70 & 53.60  & -- & 39.20 & 54.60 & 70.70  & -- & -- & -- & --  \\ 
% Diff-DR \citeyearpar{chen2025llm}     & 46.60 & 36.30 & 52.20 & 66.60  & 36.60 & 39.60 & 39.20 & 54.90  & 58.90 & 50.50 & 64.80 & 75.30  & 30.70 & 22.50 & 33.40 & 42.60  \\ \midrule
\textbf{NADEx}      & \textbf{49.03}       & \textbf{39.45}      & \textbf{57.48}      & \textbf{70.55}       & \textbf{36.84}       & \textbf{27.58}       & \textbf{45.12}      & \textbf{60.58}       & 52.17       & \textbf{43.38}      & \textbf{60.75}      & 71.93       & \underline{23.67}      & \textbf{16.10}      & \textbf{28.35}      & \textbf{43.05}       \\ \bottomrule
\end{tabular}\label{main}
\end{table*}

\subsection{Inference}
During the inference phase, the denoising network $f_\theta(*)$ is employed to reconstruct the original samples from Gaussian noise $\epsilon_m$ through a progressive refinement process. The denoising generation process of target $\mathbf{\hat{o}}_0$ is formulated as:

\begin{equation}
\footnotesize
\begin{aligned}
\mathbf{\hat{o}}_m =[\mathrm{o}_{0:m};\epsilon_m]+\mathrm{r_t+t_t}, \\
\hat{\mathrm{o}}_{0}=f_{\theta}(\mathbf{\hat{o}}_m,m).
\end{aligned}\end{equation}
Subsequently, we compute the similarity between $\hat{\mathrm{o}}_{0}$ and the object embedding matrix $\mathbf{E}_o\in \mathbb{R}^{|\mathcal{E}|\times h}$ to derive ranking results:
\begin{equation}
\footnotesize
P = \hat{\mathrm{o}}_{0} \cdot \mathbf{E}_o^\top.
\end{equation}
Entities are then ranked in descending order based on these similarity scores $P$. The top-K ranked entities are evaluated using standard metrics such as Mean Reciprocal Rank (MRR) and Hit@K.

\section{Experiments}

\subsection{Experimental Setups}

\noindent\textbf{Datasets.} Our experiments employ four benchmark datasets, including ICEWS14, ICEWS05-15, ICEWS18, and GDELT, to evaluate the proposed model. Specifically, the ICEWS datasets originate from the Integrated Crisis Early Warning System \cite{Boschee15ICEWS}, while the GDELT dataset is sourced from the Global Database of Events, Language, and Tone \cite{Leetaru13gdelt}. The data statistics are summarized in Appendix \ref{data stat}.

\noindent\textbf{Baseline Models.} We benchmark NADEx against a broad set of state-of-the-art knowledge-graph reasoning methods across three categories: \textbf{Static methods}: DisMult \cite{yang2015embedding}, ConvE \cite{dettmers2018convolutional}, RotatE\cite{sun2018rotate}; \textbf{Interpolation methods}: TTransE \cite{leblay2018deriving}, TA-DistMult \cite{garcia2018learning}, DE-SimplE \cite{goel2020diachronic}; \textbf{Extrapolation methods}: RE-NET \cite{jin2020recurrent}, Re-GCN \cite{Li21Temporal}, TITer \cite{sun2021timetraveler}, CEN \cite{li2022complex}, TiRGN \cite{li2022tirgn}, HisMatch \cite{li2022hismatch}, RETIA \cite{liu2023retia}, CENET \cite{xu2023temporal}, CRAFT \cite{zhang2024modeling}, THCN \cite{chen2024thcn}, DiffuTKG \cite{cai-etal-2024-predicting}, LogiQ \cite{chen2025enhancing}, CognTKE \cite{chen2025cogntke}. We provide baseline descriptions in Appendix \ref{baseline}. 

% The extrapolation methods are further categorized based on methodological paradigms: representation-based models: RE-NET \cite{jin2020recurrent}, CyGNet \cite{zhu2021learning}, TiRGN \cite{li2022tirgn}, RETIA \cite{liu2023retia}, CRAFT \cite{zhang2024modeling} DiffuTKG \cite{cai-etal-2024-predicting}, THCN \cite{chen2024thcn}; graph neural network-based frameworks: RE-GCN \cite{Li21Temporal}, CEN \cite{li2022complex}, HisMatch \cite{li2022hismatch}; contrastive-learning models: CENET \cite{xu2023temporal}; and rule-based or logic-guided methods: CRAFT \cite{zhang2024modeling}, LogiQ \cite{chen2025enhancing}, CognTKE \cite{chen2025cogntke}. 

\noindent\textbf{Evaluation Metrics.} To measure temporal extrapolation performance, we cast the task as masked entity prediction, where either the subject or object is held out in quadruples of the form $(s,r,?,t)$ or $(?,r,o,t)$. Predictions are scored and ranked, and we report Mean Reciprocal Rank (MRR) alongside Hits@1, Hits@3, and Hits@10. All results are computed under the time-aware filtering protocol.

\noindent\textbf{Implementation Details.} We train all models with the Adam optimizer. For ICEWS14 and ICEWS18 we use a learning rate of 1e${-3}$ and 1e${-5}$ for ICEWS05-15 and GDELT. Each model is trained for 100 epochs with an embedding dimension of 200 for both entities and relations and a dropout rate of 0.2 applied throughout. All experiments are executed on a single NVIDIA A100 (80G) GPU. All evaluations follow the filtered ranking protocol. Statistical significance is assessed using a bootstrap paired \textit{t-test}. In practice, our framework can be effectively trained within 12 hours on a single GPU on the largest dataset GDELT.

\subsection{Overall Performance}
Table \ref{main} summarizes NADEx's performance against state‑of‑the‑art (SOTA) baselines across four benchmark datasets. From these results, we make the following key observations:
\begin{itemize}[leftmargin=*]
    \item \textbf{Extrapolation‑oriented methods uniformly outperform both interpolation and time‑agnostic (“static”) approaches.} Static models, by design, ignore any temporal ordering and therefore cannot track how entity–relation interactions evolve over time. Interpolation models improve upon static baselines by filling in gaps within the observed timeline, but their training objective and architecture remain tethered to historical fact completion rather than forward‑looking prediction. In contrast, extrapolation models explicitly learn the dynamics of temporal progression, enabling them to predict future facts.

    \item \textbf{Across all four benchmarks, NADEx consistently surpasses SOTA methods on nearly every metric.} In terms of MRR, NADEx improves over the runner‑up by 3.05\% on ICEWS14 and 3.33\% on ICEWS18 and falls sligtly short on ICEWS05-15 and GDELT. Moreover, NADEx delivers substantial uplifts in Hit@K on GDELT dataset, recording performance gain of 4.68\% at Hit@1, 8.08\% at Hit@3, and 7.22\% at Hit@10.

    \item \textbf{NADEx achieves superior results on all four benchmarks compared against the leading diffusion‑based method DiffuTKG}. Notably, the performance gap widens on the more complex ICEWS05‑15 and GDELT datasets. NADEx improves MRR by 6.53\% on ICEWS05‑15 and 8.23\% on GDELT. We attribute this pattern to the richer, longer‑range temporal interactions present in ICEWS05‑15 and GDELT, which pose greater challenges for diffusion pipelines. By incorporating negative‑aware conditioning and a likelihood‑driven training objective, NADEx more effectively models these intricate dynamics, whereas DiffuTKG's reconstruction loss struggles to discriminate plausible from spurious future facts under such complex scenarios.

\end{itemize}

\begin{table}[t]
\setlength{\tabcolsep}{3.3pt} % 减小列间距
\fontsize{8pt}{10pt}\selectfont % 设置特定字号和行距
\caption{Ablation study results ICEWS14 and GDELT datasets in terms of MRR and Hit@1/10.}
\begin{tabular}{@{}ccccccc@{}}
\toprule
\multirow{2}{*}{Settings} & \multicolumn{3}{c}{ICEWS14} & \multicolumn{3}{c}{GDELT} \\ \cmidrule(l){2-7} 
 & MRR   & Hit@1   & Hit@10   & MRR   & Hit@1   & Hit@10  \\ \midrule
NADEx       & \textbf{49.03}      & \textbf{39.45}        & \textbf{70.55} & \textbf{23.67}      & \textbf{16.10}        & \textbf{43.05}        \\ \midrule
w/o. $E_\Delta$  & 43.16      & 33.35        & 62.06 &  18.82     &   12.26      &   36.32      \\
w/o. $E_r$  & 40.05      & 30.23        & 58.47 &   18.21    &    11.76     &   35.52      \\
w/o. $E_r \& E_\Delta$  & 32.48      & 22.71        & 51.69 &    18.06   &  11.66       &     35.21    \\
w/o. neg   & 47.42      & 36.39        & 67.61 & 21.25      & 14.22        & 42.34        \\ \bottomrule
\end{tabular}\label{ablation}
\end{table}

\subsection{Ablation Studies}
To quantify the contribution of each NADEx component, we perform ablations on ICEWS14 and GDELT, measuring MRR, Hit@1, and Hit@10, as presented in Table \ref{ablation}. Removing the time‐interval embedding (“w/o $E_{\Delta}$”) degrades MRR by 5.87 points on ICEWS14 and 4.85 points on GDELT, highlighting the importance of explicit temporal cues. Omitting the relation‐type embedding (“w/o $E_{r}$”) incurs an even larger drop across all metrics, confirming that relational context is essential for accurate reconstruction. Omitting both temporal and relational embeddings simultaneously (“w/o $E_{r} \& E_{\Delta}$”) Omitting both temporal and relational embeddings simultaneously. Lastly, eliminating our negative-sampling strategy (“w/o neg”) consistently lowers performance, affirming the value of negative-aware training in improving the model's discriminative capability between plausible and implausible event predictions. Together, these results confirm that NADEx's strength derives not merely from its generative backbone but from the harmonious integration of temporal, relational, and negative‐aware conditioning.

\begin{table}[t]
\caption{Performance of predicting unseen events in terms of MRR and Hit@1 on ICEWS14 and ICEWS18.}
\setlength{\tabcolsep}{12pt} % 减小列间距
\fontsize{8pt}{8pt}\selectfont % 设置特定字号和行距
\begin{tabular}{@{}ccccc@{}}
\toprule
\multirow{2}{*}{Models} & \multicolumn{2}{c}{ICEWS14} & \multicolumn{2}{c}{ICEWS18} \\ \cmidrule(l){2-5} 
  & MRR  & Hit@1        & MRR  & Hit@1        \\ \midrule
RE-GCN    & 23.26        & 13.91        & 15.08        & 7.09 \\
CEN       & 22.06        & 13.28        & 15.41        & 8.20 \\
RETIA     & 24.17        & 14.67        & 16.62        & 9.08 \\
HisMatch  & 27.49        & \underline{19.04}        & 17.51        & 11.13        \\
DiffuTKG  & 25.22        & 15.23        & 16.48        & 8.84 \\
LogCL     & \underline{29.19}        & 18.72        & \underline{18.40}        & \underline{11.74}        \\ \midrule
NADEx     & \textbf{30.85}        & \textbf{20.08}        & \textbf{19.52}        & \textbf{12.17}        \\
\textit{Improve.}     & \textit{5.89\%}        & \textit{5.46\%}        & \textit{6.09\%}        & \textit{3.66\%}        \\\bottomrule
\end{tabular}\label{unseen}
\end{table}

\begin{figure}[t]
	\centering
	\subfloat[Performance of MRR]{
		\includegraphics[scale=0.15]{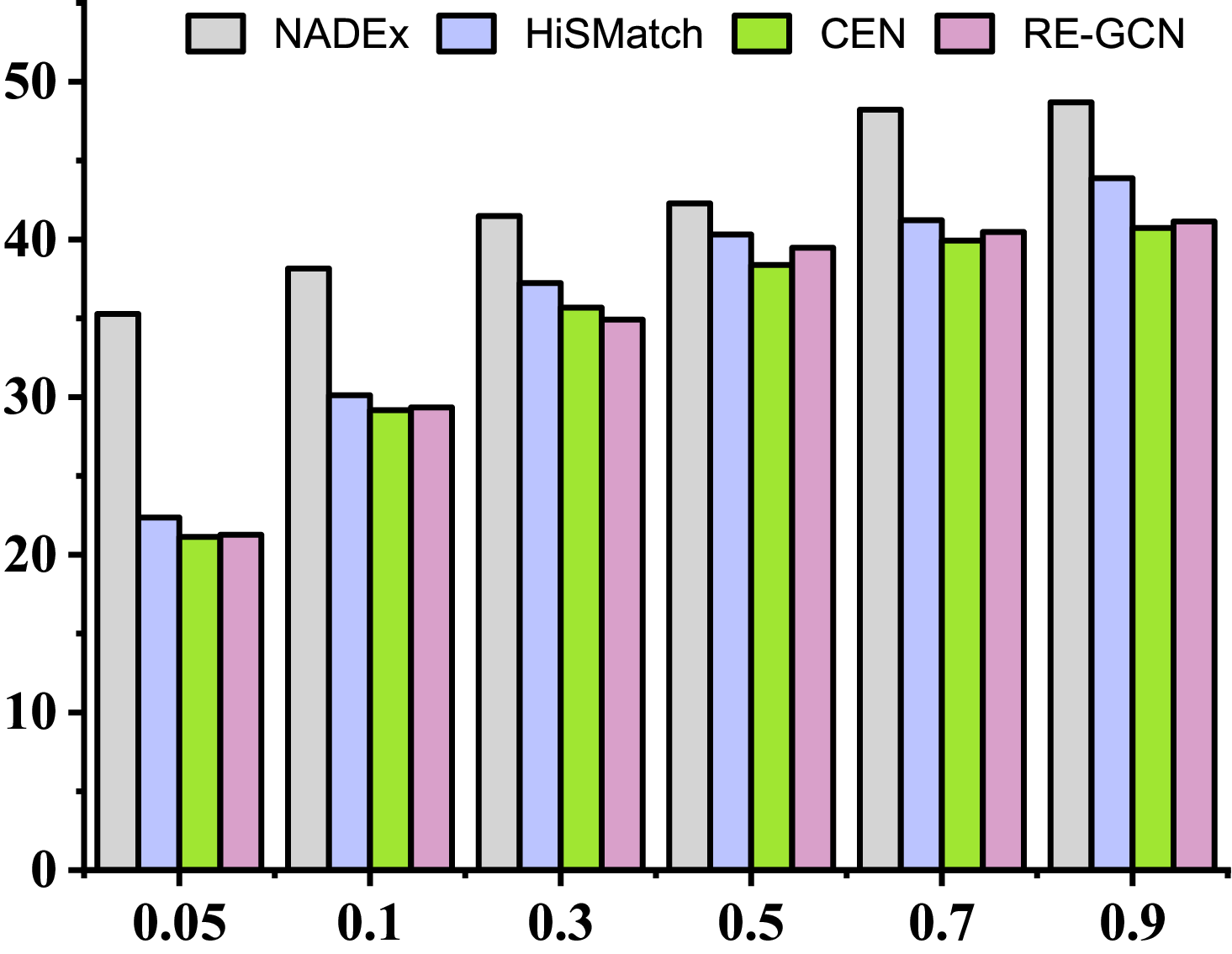}
	}%
	\subfloat[Performance of Hit@1]{		
		\includegraphics[scale=0.15]{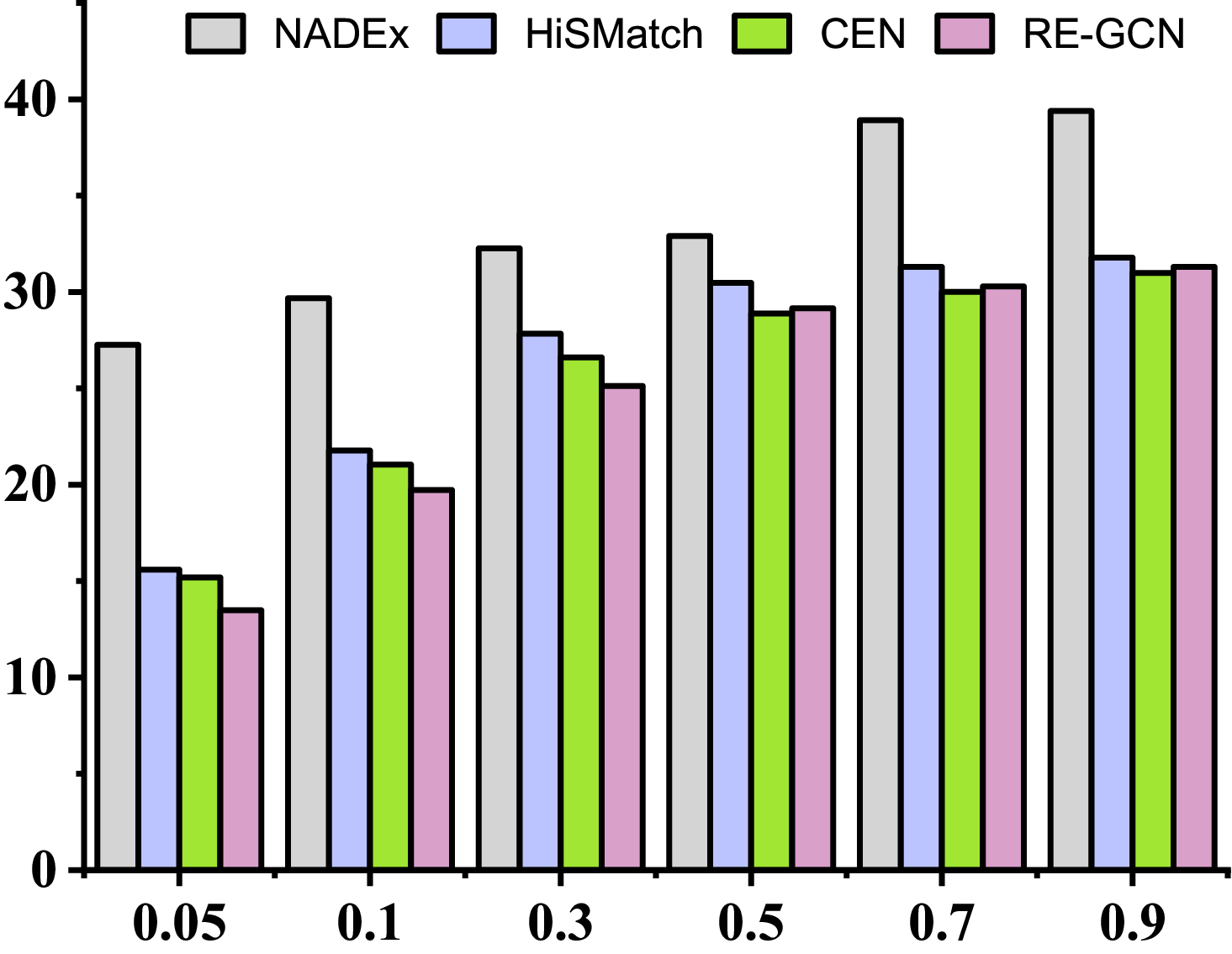}
	}%
    
	% \subfloat[Hit@3]{
	% 	\includegraphics[scale=0.2]{HIT3.eps}
	% }%
	% \subfloat[Hit@10]{		
	% 	\includegraphics[scale=0.2]{HIT10.eps}
	% }%
	\centering
	\caption{Performance of NADEx compared with three baselines in terms of MRR and Hit@1 on ICEWS14 under different training data scale settings.}
	\label{fig5.5}
\end{figure}

\begin{table*}[t]
\caption{Case studies on ICEWS14 showing each model's top-5 predicted entities for sample queries.}
\setlength{\tabcolsep}{7pt} % 减小列间距
\fontsize{8pt}{10pt}\selectfont % 设置特定字号和行距
\begin{tabular}{@{}ccccc@{}}
\toprule
Query  & \textbf{NADEx}        & LogCL     & DiffuTKG      & THCN     \\ \midrule
\begin{tabular}[c]{@{}c@{}}<China, 057, ?>\\ Date: 2014-12-3\\ \textbf{Label: Maldives}\end{tabular}    & \begin{tabular}[c]{@{}c@{}}\textcolor{red}{Maldives, 0.534}\\ Taavi Roivas, 0.167\\ Angola, 0.019\\ South Korea, 0.010\\ Malaysia, 0.006\end{tabular} & \begin{tabular}[c]{@{}c@{}}Japan, 0.457\\ South Korea, 0.195\\ Thailand, 0.081\\ Legislature, 0.043\\ Malaysia 0.035\end{tabular} & \begin{tabular}[c]{@{}c@{}}Laos, 0.324\\ Thailand, 0.212\\ Japan, 0.073\\ Iran, 0.053\\ Malaysia, 0.043\end{tabular}  & \begin{tabular}[c]{@{}c@{}}Laos, 0.298\\ Malaysia, 0.141\\ Iran, 0.102\\ Thailand, 0.095\\ Japan, 0.074\end{tabular} \\ \midrule
\begin{tabular}[c]{@{}c@{}}<South Korea, 060, ?>\\ Date: 2014-12-9\\ \textbf{Label: Japan}\end{tabular} & \begin{tabular}[c]{@{}c@{}}\textcolor{red}{Japan, 0.414}\\ North Korea, 0.347\\ China, 0.084\\ Philippines, 0.028\\ Iran, 0.026\end{tabular}        & \begin{tabular}[c]{@{}c@{}}\textcolor{red}{Japan, 0.289}\\ North Korea, 0.128\\ North Korea, 0.027\\ Japan (Newspaper), 0.021\\ China, 0.006\end{tabular} & \begin{tabular}[c]{@{}c@{}}North Korea (Military), 0.128\\ China, 0.122\\ Qatar, 0.078\\ Iran, 0.060\\ \textcolor{red}{Japan, 0.025}\end{tabular}  & \begin{tabular}[c]{@{}c@{}}North Korea (Military), 0.329\\ North Korea, 0.122\\ China, 0.113\\ Iran, 0.061\\ South Korea (Police), 0.025\end{tabular}\\ \bottomrule
\end{tabular}\label{case}
\end{table*}

\subsection{Effectiveness of Unseen Events}
To further verify the capacity of NADEx in capturing discriminative and generalizable representations for unseen event queries, we specifically evaluate its predictive performance on ICEWS14 and ICEWS18 datasets, both characterized by a relatively higher proportion of previously unobserved events. As reported in Table \ref{unseen}, NADEx consistently surpasses existing state-of-the-art methods by significant margins. Specifically, NADEx achieves remarkable improvements of approximately 4.43\% and 5.40\% in terms of absolute MRR on ICEWS14 and ICEWS18, respectively, compared with the strongest baseline (LogCL). Additionally, NADEx attains notable enhancements of 3.10\% and 3.43\% in absolute Hit@1 performance on the two datasets, underscoring its superior precision in accurately identifying true future entities. These empirical results substantiate that explicitly incorporating negative contexts and employing a diffusion-based generative framework effectively enhances the model's discriminative power, enabling robust extrapolation to rare/unseen future events under uncertainty.

\subsection{Performance under Limited Data}
To assess data-efficiency, we subsample the ICEWS14 training set at several fractions (5\%, 10\%, 30\%, 50\%, 70\%, and 90\%) while keeping the validation and test splits intact. Figure \ref{fig5.5} shows that every model deteriorates as the available data diminish, reflecting the increased difficulty of learning reliable representations under sparse supervision. Even so, NADEx maintains a clear lead over HisMatch, CEN, and RE-GCN in both MRR and Hit@1 at every sampling ratio. Notably, between the 50\% and 5\% splits, NADEx's MRR degrades by only 7\%, whereas all baselines suffer losses exceeding 15\%. A similar trend holds for Hit@1. We attribute this robustness to the negative-aware diffusion mechanism, which supplies informative counter-examples during generation, the model learns fine-grained decision boundaries with far fewer labeled instances, and it maintains calibrated uncertainty estimates that remain reliable even when supervision is extremely sparse. Notably, the performance margin in favor of NADEx widens as the training fraction decreases, underscoring its superior generalization in low-resource settings. We attribute these strengths to NADEx's negative-aware diffusion mechanism: by injecting negatives drawn from each mini-batch into the generation process, the model learns more discriminative decision boundaries with fewer samples.

\subsection{Case Study}
Table \ref{case} presents two illustrative queries on ICEWS14, comparing NADEx against LogCL and DiffuTKG by listing each model's top-5 predicted entities and their associated scores. For the first query <China, 057, ?, 2014-12-3>, NADEx places the correct target, \textbf{Maldives} (score 0.534), firmly at the top, followed by semantically coherent but less plausible alternatives such as Taavi Roivas (0.167) and Angola (0.019). LogCL instead places Japan first (0.457) and omits Maldives from its top-5; DiffuTKG similarly favors Laos (0.324) and does not rank Maldives among its candidates. For the second query <South Korea, 060, ?, 2014-12-9>, NADEx correctly identifies \textbf{Japan} as the most likely future partner (0.414), with close runner-up North Korea (0.347) reflecting plausible regional dynamics. LogCL also selects Japan but with lower confidence (0.289) and a more diffuse allocation across other entities; DiffuTKG ranks North Korea highest (0.128) and places Japan fifth (0.025). Across both queries, NADEx concentrates probability mass on the true target and suppresses near-neighbor distractors. The negative-aware conditioning sharpens the posterior and yields more calibrated confidence relative to the baselines.

\begin{table}[]
\caption{Computational efficiency test on ICEWS14/18.}
\setlength{\tabcolsep}{6pt} % 减小列间距
\fontsize{10pt}{12pt}\selectfont % 设置特定字号和行距
\begin{tabular}{@{}ccccc@{}}
\toprule
\multirow{2}{*}{Model} & \multicolumn{2}{c}{ICEWS14} & \multicolumn{2}{c}{ICEWS18} \\ \cmidrule(l){2-5} 
 & \begin{tabular}[c]{@{}c@{}}Inf.\\ Time\end{tabular} & Params. & \begin{tabular}[c]{@{}c@{}}Inf.\\ Time\end{tabular} & Params. \\ \midrule
RE-GCN   & 28.96s & 26.52Mb & 393.67s & 42.68Mb \\
TiRGN    & 32.88s & 43.35Mb & 159.90s & 59.59Mb \\
DiffuTKG & 13.49s & 18.35Mb & 135.80s & 34.46Mb \\ \midrule
NADEx    & \textbf{10.91s}       & \textbf{16.30Mb}    & \textbf{96.95s}      & \textbf{32.42Mb}    \\ \bottomrule
\end{tabular}\label{comp}
\end{table}

\begin{table*}[t]
\caption{Performance comparison (\%) with LLM-driven approaches on ICEWS14 and ICEWS18. The best results are highlighted in \textbf{bold}, while the second-best results are \underline{underlined}. The experimental results are retrieved from \cite{luo2024chain} and \cite{chen2025llm}, respectively.}
\setlength{\tabcolsep}{2.2pt} % 减小列间距
\fontsize{10pt}{10pt}\selectfont % 设置特定字号和行距
\resizebox{\linewidth}{!}{
\begin{tabular}{@{}cc|cccccccc@{}}
\toprule
\multicolumn{2}{c|}{\multirow{2}{*}{Models}}  & \multirow{2}{*}{\begin{tabular}[c]{@{}c@{}}GenTKG\\ \citeyearpar{liao2024gentkg}\end{tabular}} & \multirow{2}{*}{\begin{tabular}[c]{@{}c@{}}GPT-NeoX\\ \citeyearpar{black2022gpt}\end{tabular}} & \multirow{2}{*}{\begin{tabular}[c]{@{}c@{}}Llama-ICL\\ \citeyearpar{touvronllama}\end{tabular}} & \multirow{2}{*}{\begin{tabular}[c]{@{}c@{}}Vicuna-ICL\\ \citeyearpar{chiang2023vicuna}\end{tabular}} & \multirow{2}{*}{\begin{tabular}[c]{@{}c@{}}Llama-CoH\\ \citeyearpar{luo2024chain}\end{tabular}} & \multirow{2}{*}{\begin{tabular}[c]{@{}c@{}}Vicuna-CoH\\ \citeyearpar{luo2024chain}\end{tabular}} & \multirow{2}{*}{\begin{tabular}[c]{@{}c@{}}LLM-DR\\ \citeyearpar{chen2025llm}\end{tabular}} & \multirow{2}{*}{\textbf{NADEx}} \\
\multicolumn{2}{c|}{}     & & &       &      &       &       & &  \\ \midrule
\multicolumn{1}{c|}{\multirow{3}{*}{ICEWS14}} & Hit@1  & 34.90       & 29.50 & 28.60 & 28.10  & 34.90 & 32.80 & \textbf{40.60}       & \underline{39.45}    \\
\multicolumn{1}{c|}{} & Hit@3  & 47.30       & 40.60 & 39.70 & 39.10  & 47.00 & 45.70 & \underline{55.80}       & \textbf{57.48} \\
\multicolumn{1}{c|}{} & Hit@10 & 61.90       & 47.50 & 45.30 & 45.30  & 59.10 & 65.60 & \underline{67.00}       & \textbf{70.55} \\ \midrule
\multicolumn{1}{c|}{\multirow{3}{*}{ICEWS18}} & Hit@1  & 21.50       & 17.70 & 17.70 & 17.20  & 22.30 & 20.90 & \textbf{30.40}  & \underline{27.58}    \\
\multicolumn{1}{c|}{} & Hit@3  & 36.60       & 29.00 & 29.40 & 28.80  & 36.30 & 34.70 & \underline{40.90}       & \textbf{45.12} \\
\multicolumn{1}{c|}{} & Hit@10 & 49.60       & 38.20 & 36.40 & 36.40  & 52.20 & 53.60 & \underline{55.60}       & \textbf{60.58} \\ \midrule
\multicolumn{2}{c|}{AVG.} & 41.97       & 35.42 & 32.85 & 32.48  & 41.97 & 42.22 & \underline{48.38}       & \textbf{50.13} \\ \bottomrule
\end{tabular}}
\label{llm}
\end{table*}

\subsection{Computational Efficiency}
On computational efficiency, NADEx attains the best speed–size trade-off on both ICEWS14 and ICEWS18, as shown in Figure \ref{comp}. It yields the lowest inference time and smallest parameter 10.91s/16.30Mb on ICEWS14 and 96.95s/32.42Mb on ICEWS18. Relative to the diffusion baseline DiffuTKG, NADEx reduces inference time by 19.1\% on ICEWS14 and 28.6\% on ICEWS18 with smaller parameter sizes. Against graph-based baselines, NADEx is much faster than RE-GCN/TiRGN on ICEWS14 ICEWS18, while using 20\% to 60\% fewer parameters. When scaling from ICEWS14 to ICEWS18, the inference time of NADEx grows by 8.89×, compared with 10.07× for DiffuTKG and 13.59× for RE-GCN, indicating more favorable scaling with dataset size. These results demonstrate that NADEx achieves better computational efficiency without increasing model capacity.

\subsection{Comparison with LLM-based Approaches}
In this section, we systematically evaluate our proposed approach against recent state-of-the-art large language model (LLM)-driven temporal knowledge graph (TKG) reasoning methods. Specifically, we benchmark our model against prominent LLM-based baselines including GenTKG \cite{liao2024gentkg}, GPT-NeoX \cite{lee2023temporal}, Llama-2 \cite{luo2024chain}, Vicuna \cite{luo2024chain}, and LLM-DR \cite{wang2024large} on two benchmarks, ICEWS14 and ICEWS18. Performance is measured across standard ranking metrics: Hit@1, Hit@3, and Hit@10, and Table \ref{llm} summarizes the comparative results. Experimental results indicate that certain LLM-based approaches such as GenTKG and GPT-NeoX do not consistently outperform traditional TKG reasoning methods. This observation can be primarily attributed to the overly general logical rules derived from LLM-generated reasoning chains, which may fail to precisely model the intricate temporal dynamics inherent in TKGs. Notably, LLM‑DR, which continuously refines its rule set to mirror the empirical distribution, yields the best Hit@1 scores (40.60\% on ICEWS14 and 30.40\% on ICEWS18 datasets). However, its Hit@3 and Hit@10 scores decline because the iterative rule‑pruning aggressively narrows each query to a high‑precision candidate set, increasing the likelihood that the top prediction is correct while simultaneously sacrificing lower‑rank diversity and overall recall. In contrast, our proposed approach achieves robust performance across all metrics, surpassing all baseline methods on Hit@3 and Hit@10, and obtaining a strong Hit@1 result second only to LLM-DR. With an average overall hit score of 50.13\% across both datasets.

\section{Conclusion}
In this paper, we propose NADEx, a novel Negative-Aware Diffusion model specifically designed for TKG extrapolation. By recasting TKG reasoning as a sequence‐denoising problem, NADEx perturbs target‐entity embeddings with Gaussian noise and then leverages a Transformer‐based denoiser to reconstruct the true entity representation. Crucially, we embed batch‐wise negative prototypes into the diffusion process and train under objectives that combine cross-entropy reconstruction with a cosine-alignment regularizer. This formulation not only sharpens the model's ability to discriminate between plausible and implausible future events but also preserves the generative strengths of diffusion dynamics. Comprehensive experiments conducted across four widely-used TKG benchmarks demonstrate that NADEx consistently achieves state-of-the-art performance.

\section*{Acknowledgments}
We thank the anonymous reviewers for their valuable discussion and feedback. This work was supported by the National Natural Science Foundation of China (U22B2061), and the National Key R\&D Program of China (2022YFB4300603).

\section*{Limitation}
One limitation of this work lies in its reliance on four widely used TKG benchmarks (ICEWS14, ICEWS05–15, ICEWS18, and GDELT), although standard, represent a relatively narrow spectrum of event data. These datasets are dominated by political and international relations events, with limited diversity in domains such as science, economics, or natural disasters. As a result, the model’s generalizability to other types of temporal knowledge graphs remains untested. In addition, while the proposed negative-aware diffusion framework introduces a cosine-alignment regularizer to incorporate negative prototypes, the construction of negatives is limited to batch-level sampling. This strategy may not fully capture semantically hard negatives or long-tail entities, potentially leading to overly coarse approximations of implausible events.

\section*{Ethics Statement}
This study adheres to established ethical standards. We rely solely on datasets that were previously collected and annotated by earlier research, and our use of these resources entails no new data gathering, no interaction with human subjects, and no processing of personally identifiable or otherwise private information. The models and results are presented for scientific analysis and benchmarking and are not intended for surveillance, deception, or any form of harm. In all cases, we respect data-use terms and privacy safeguards and ensure that our work does not endanger the rights, safety, or dignity of any individual or group.
% This document has been adapted
% by Steven Bethard, Ryan Cotterell and Rui Yan
% from the instructions for earlier ACL and NAACL proceedings, including those for
% ACL 2019 by Douwe Kiela and Ivan Vuli\'{c},
% NAACL 2019 by Stephanie Lukin and Alla Roskovskaya,
% ACL 2018 by Shay Cohen, Kevin Gimpel, and Wei Lu,
% NAACL 2018 by Margaret Mitchell and Stephanie Lukin,
% Bib\TeX{} suggestions for (NA)ACL 2017/2018 from Jason Eisner,
% ACL 2017 by Dan Gildea and Min-Yen Kan,
% NAACL 2017 by Margaret Mitchell,
% ACL 2012 by Maggie Li and Michael White,
% ACL 2010 by Jing-Shin Chang and Philipp Koehn,
% ACL 2008 by Johanna D. Moore, Simone Teufel, James Allan, and Sadaoki Furui,
% ACL 2005 by Hwee Tou Ng and Kemal Oflazer,
% ACL 2002 by Eugene Charniak and Dekang Lin,
% and earlier ACL and EACL formats written by several people, including
% John Chen, Henry S. Thompson and Donald Walker.
% Additional elements were taken from the formatting instructions of the \emph{International Joint Conference on Artificial Intelligence} and the \emph{Conference on Computer Vision and Pattern Recognition}.

% Bibliography entries for the entire Anthology, followed by custom entries
%\bibliography{anthology,custom}
% Custom bibliography entries only
\bibliography{custom}

@data{Boschee15ICEWS,
author = {Boschee, Elizabeth and Lautenschlager, Jennifer and O'Brien, Sean and Shellman, Steve and Starz, James and Ward, Michael},
publisher = {Harvard Dataverse},
title = {{ICEWS Coded Event Data}},
UNF = {UNF:6:NOSHB7wyt0SQ8sMg7+w38w==},
year = {2015},
version = {V37},
doi = {10.7910/DVN/28075},
url = {https://doi.org/10.7910/DVN/28075}
}

@article{Leetaru13gdelt,
  author = {Leetaru, Kalev and Schrodt, Philip A.},
  description = {CiteSeerX — GDELT: Global data on events, location, and tone},
  journal = {ISA Annual Convention},
  title = {GDELT: Global data on events, location, and tone},
 pages = {1–49},
volume = {2},
  url = {http://citeseerx.ist.psu.edu/viewdoc/summary?doi=10.1.1.686.6605},
  year = 2013
}

@inproceedings{Li21Temporal,
author = {Li, Zixuan and Jin, Xiaolong and Li, Wei and Guan, Saiping and Guo, Jiafeng and Shen, Huawei and Wang, Yuanzhuo and Cheng, Xueqi},
title = {Temporal Knowledge Graph Reasoning Based on Evolutional Representation Learning},
year = {2021},
isbn = {9781450380379},
publisher = {Association for Computing Machinery},
address = {New York, NY, USA},
url = {https://doi.org/10.1145/3404835.3462963},
doi = {10.1145/3404835.3462963},
booktitle = {Proceedings of the 44th International ACM SIGIR Conference on Research and Development in Information Retrieval},
pages = {408–417},
numpages = {10},
location = {Virtual Event, Canada},
series = {SIGIR '21}
}

@inproceedings{cai-etal-2024-predicting,
    title = "Predicting the Unpredictable: Uncertainty-Aware Reasoning over Temporal Knowledge Graphs via Diffusion Process",
    author = "Cai, Yuxiang  and
      Liu, Qiao  and
      Gan, Yanglei  and
      Li, Changlin  and
      Liu, Xueyi  and
      Lin, Run  and
      Luo, Da  and 
      Jiaye Yang",
    booktitle = "Findings of the Association for Computational Linguistics: ACL 2024",
    month = aug,
    year = "2024",
    address = "Bangkok, Thailand",
    publisher = "Association for Computational Linguistics",
    url = "https://aclanthology.org/2024.findings-acl.343/",
    doi = "10.18653/v1/2024.findings-acl.343",
    pages = "5766--5778"
}

@inproceedings{chen2024local,
  title={Local-global history-aware contrastive learning for temporal knowledge graph reasoning},
  author={Chen, Wei and Wan, Huaiyu and Wu, Yuting and Zhao, Shuyuan and Cheng, Jiayaqi and Li, Yuxin and Lin, Youfang},
  booktitle={2024 IEEE 40th International Conference on Data Engineering (ICDE)},
  pages={733--746},
  year={2024},
  organization={IEEE}
}

@inproceedings{chen2025llm,
  title={LLM-DR: A Novel LLM-Aided Diffusion Model for Rule Generation on Temporal Knowledge Graphs},
  author={Chen, Kai and Song, Xin and Wang, Ye and Gao, Liqun and Li, Aiping and Zhao, Xiaojuan and Zhou, Bin and Xie, Yalong},
  booktitle={Proceedings of the AAAI Conference on Artificial Intelligence},
  volume={39},
  number={11},
  pages={11481--11489},
  year={2025}
}

@inproceedings{chen2025enhancing,
  title={Enhancing Extrapolation Reasoning on Temporal Knowledge Graphs with Logic Rules and Queries},
  author={Chen, Tingxuan and Yang, Liu and Wang, Zidong and Luo, Shuai and Long, Jun},
  booktitle={ICASSP 2025-2025 IEEE International Conference on Acoustics, Speech and Signal Processing (ICASSP)},
  pages={1--5},
  year={2025},
  organization={IEEE}
}

@inproceedings{chen2025cogntke,
  title={CognTKE: A Cognitive Temporal Knowledge Extrapolation Framework},
  author={Chen, Wei and Wu, Yuting and Wu, Shuhan and Zhang, Zhiyu and Liao, Mengqi and Lin, Youfang and Wan, Huaiyu},
  booktitle={Proceedings of the AAAI Conference on Artificial Intelligence},
  volume={39},
  number={14},
  pages={14815--14823},
  year={2025}
}

@inproceedings{garcia2018learning,
  title={Learning Sequence Encoders for Temporal Knowledge Graph Completion},
  author={Garcia-Duran, Alberto and Duman{\v{c}}i{\'c}, Sebastijan and Niepert, Mathias},
  booktitle={Proceedings of the 2018 Conference on Empirical Methods in Natural Language Processing},
  pages={4816--4821},
  year={2018}
}

@inproceedings{leblay2018deriving,
  title={Deriving validity time in knowledge graph},
  author={Leblay, Julien and Chekol, Melisachew Wudage},
  booktitle={Companion proceedings of the the web conference 2018},
  pages={1771--1776},
  year={2018}
}

@inproceedings{yang2015embedding,
  title={Embedding Entities and Relations for Learning and Inference in Knowledge Bases},
  author={Yang, Bishan and Yih, Scott Wen-tau and He, Xiaodong and Gao, Jianfeng and Deng, Li},
  booktitle={Proceedings of the International Conference on Learning Representations (ICLR) 2015},
  year={2015}
}

@inproceedings{dettmers2018convolutional,
  title={Convolutional 2d knowledge graph embeddings},
  author={Dettmers, Tim and Minervini, Pasquale and Stenetorp, Pontus and Riedel, Sebastian},
  booktitle={Proceedings of the AAAI conference on artificial intelligence},
  volume={32},
  number={1},
  year={2018}
}

@inproceedings{sun2018rotate,
title={RotatE: Knowledge Graph Embedding by Relational Rotation in Complex Space},
author={Zhiqing Sun and Zhi-Hong Deng and Jian-Yun Nie and Jian Tang},
booktitle={International Conference on Learning Representations},
year={2019},
url={https://openreview.net/forum?id=HkgEQnRqYQ},
}

@inproceedings{goel2020diachronic,
  title={Diachronic embedding for temporal knowledge graph completion},
  author={Goel, Rishab and Kazemi, Seyed Mehran and Brubaker, Marcus and Poupart, Pascal},
  booktitle={Proceedings of the AAAI conference on artificial intelligence},
  volume={34},
  number={04},
  pages={3988--3995},
  year={2020}
}

@inproceedings{jin2020recurrent,
  title={Recurrent Event Network: Autoregressive Structure Inferenceover Temporal Knowledge Graphs},
  author={Jin, Woojeong and Qu, Meng and Jin, Xisen and Ren, Xiang},
  booktitle={Proceedings of the 2020 Conference on Empirical Methods in Natural Language Processing (EMNLP)},
  pages={6669--6683},
  year={2020}
}

@inproceedings{zhu2021learning,
  title={Learning from history: Modeling temporal knowledge graphs with sequential copy-generation networks},
  author={Zhu, Cunchao and Chen, Muhao and Fan, Changjun and Cheng, Guangquan and Zhang, Yan},
  booktitle={Proceedings of the AAAI conference on artificial intelligence},
  volume={35},
  number={5},
  pages={4732--4740},
  year={2021}
}

@inproceedings{li2022complex,
  title={Complex Evolutional Pattern Learning for Temporal Knowledge Graph Reasoning},
  author={Li, Zixuan and Guan, Saiping and Jin, Xiaolong and Peng, Weihua and Lyu, Yajuan and Zhu, Yong and Bai, Long and Li, Wei and Guo, Jiafeng and Cheng, Xueqi},
  booktitle={Proceedings of the 60th Annual Meeting of the Association for Computational Linguistics (Volume 2: Short Papers)},
  pages={290--296},
  year={2022}
}

@inproceedings{li2022tirgn,
  title={TiRGN: Time-Guided Recurrent Graph Network with Local-Global Historical Patterns for Temporal Knowledge Graph Reasoning.},
  author={Li, Yujia and Sun, Shiliang and Zhao, Jing},
  booktitle={IJCAI},
  pages={2152--2158},
  year={2022}
}

@inproceedings{li2022hismatch,
  title={HiSMatch: Historical Structure Matching based Temporal Knowledge Graph Reasoning},
  author={Li, Zixuan and Hou, Zhongni and Guan, Saiping and Jin, Xiaolong and Peng, Weihua and Bai, Long and Lyu, Yajuan and Li, Wei and Guo, Jiafeng and Cheng, Xueqi},
  booktitle={Findings of the Association for Computational Linguistics: EMNLP 2022},
  pages={7328--7338},
  year={2022}
}

@inproceedings{liu2023retia,
  title={RETIA: relation-entity twin-interact aggregation for temporal knowledge graph extrapolation},
  author={Liu, Kangzheng and Zhao, Feng and Xu, Guandong and Wang, Xianzhi and Jin, Hai},
  booktitle={2023 IEEE 39th international conference on data engineering (ICDE)},
  pages={1761--1774},
  year={2023},
  organization={IEEE}
}

@inproceedings{xu2023temporal,
  title={Temporal knowledge graph reasoning with historical contrastive learning},
  author={Xu, Yi and Ou, Junjie and Xu, Hui and Fu, Luoyi},
  booktitle={Proceedings of the AAAI conference on artificial intelligence},
  volume={37},
  number={4},
  pages={4765--4773},
  year={2023}
}

@article{chen2024thcn,
  title={THCN: A Hawkes Process Based Temporal Causal Convolutional Network for Extrapolation Reasoning in Temporal Knowledge Graphs},
  author={Chen, Tingxuan and Long, Jun and Wang, Zidong and Luo, Shuai and Huang, Jincai and Yang, Liu},
  journal={IEEE Transactions on Knowledge and Data Engineering},
  year={2024},
  publisher={IEEE}
}

@inproceedings{zhang2024modeling,
  title={Modeling Historical Relevant and Local Frequency Context for Representation-Based Temporal Knowledge Graph Forecasting},
  author={Zhang, Shengzhe and Wei, Wei and Huang, Rikui and Xie, Wenfeng and Chen, Dangyang},
  booktitle={Findings of the Association for Computational Linguistics: EMNLP 2024},
  pages={7675--7686},
  year={2024}
}

@article{liang2024survey,
  title={A survey of knowledge graph reasoning on graph types: Static, dynamic, and multi-modal},
  author={Liang, Ke and Meng, Lingyuan and Liu, Meng and Liu, Yue and Tu, Wenxuan and Wang, Siwei and Zhou, Sihang and Liu, Xinwang and Sun, Fuchun and He, Kunlun},
  journal={IEEE Transactions on Pattern Analysis and Machine Intelligence},
  year={2024},
  publisher={IEEE}
}

@article{ji2021survey,
  title={A survey on knowledge graphs: Representation, acquisition, and applications},
  author={Ji, Shaoxiong and Pan, Shirui and Cambria, Erik and Marttinen, Pekka and Philip, S Yu},
  journal={IEEE transactions on neural networks and learning systems},
  volume={33},
  number={2},
  pages={494--514},
  year={2021},
  publisher={IEEE}
}

@inproceedings{cai2023temporal,
  title={Temporal knowledge graph completion: a survey},
  author={Cai, Borui and Xiang, Yong and Gao, Longxiang and Zhang, He and Li, Yunfeng and Li, Jianxin},
  booktitle={Proceedings of the Thirty-Second International Joint Conference on Artificial Intelligence},
  pages={6545--6553},
  year={2023}
}

@inproceedings{trivedi2017know,
  title={Know-evolve: Deep temporal reasoning for dynamic knowledge graphs},
  author={Trivedi, Rakshit and Dai, Hanjun and Wang, Yichen and Song, Le},
  booktitle={international conference on machine learning},
  pages={3462--3471},
  year={2017},
  organization={PMLR}
}

@article{bordes2013translating,
  title={Translating embeddings for modeling multi-relational data},
  author={Bordes, Antoine and Usunier, Nicolas and Garcia-Duran, Alberto and Weston, Jason and Yakhnenko, Oksana},
  journal={Advances in neural information processing systems},
  volume={26},
  year={2013}
}

@inproceedings{zhang2023learning,
  title={Learning long-and short-term representations for temporal knowledge graph reasoning},
  author={Zhang, Mengqi and Xia, Yuwei and Liu, Qiang and Wu, Shu and Wang, Liang},
  booktitle={Proceedings of the ACM web conference 2023},
  pages={2412--2422},
  year={2023}
}

@inproceedings{sun2021timetraveler,
  title={TimeTraveler: Reinforcement Learning for Temporal Knowledge Graph Forecasting},
  author={Sun, Haohai and Zhong, Jialun and Ma, Yunpu and Han, Zhen and He, Kun},
  booktitle={Proceedings of the 2021 Conference on Empirical Methods in Natural Language Processing},
  pages={8306--8319},
  year={2021}
}

@inproceedings{zheng2023dream,
  title={DREAM: Adaptive reinforcement learning based on attention mechanism for temporal knowledge graph reasoning},
  author={Zheng, Shangfei and Yin, Hongzhi and Chen, Tong and Nguyen, Quoc Viet Hung and Chen, Wei and Zhao, Lei},
  booktitle={Proceedings of the 46th international ACM SIGIR conference on research and development in information retrieval},
  pages={1578--1588},
  year={2023}
}

@inproceedings{cao2025dpcl,
  title={DPCL-Diff: Temporal Knowledge Graph Reasoning Based on Graph Node Diffusion Model with Dual-Domain Periodic Contrastive Learning},
  author={Cao, Yukun and Wang, Lisheng and Huang, Luobin},
  booktitle={Proceedings of the AAAI Conference on Artificial Intelligence},
  volume={39},
  number={14},
  pages={14806--14814},
  year={2025}
}

@inproceedings{han2021explainable,
title={Explainable Subgraph Reasoning for Forecasting on Temporal Knowledge Graphs},
author={Zhen Han and Peng Chen and Yunpu Ma and Volker Tresp},
booktitle={International Conference on Learning Representations},
year={2021},
url={https://openreview.net/forum?id=pGIHq1m7PU}
}

@inproceedings{liu2022tlogic,
  title={Tlogic: Temporal logical rules for explainable link forecasting on temporal knowledge graphs},
  author={Liu, Yushan and Ma, Yunpu and Hildebrandt, Marcel and Joblin, Mitchell and Tresp, Volker},
  booktitle={Proceedings of the AAAI conference on artificial intelligence},
  volume={36},
  number={4},
  pages={4120--4127},
  year={2022}
}

@inproceedings{dong2023adaptive,
  title={Adaptive path-memory network for temporal knowledge graph reasoning},
  author={Dong, Hao and Ning, Zhiyuan and Wang, Pengyang and Qiao, Ziyue and Wang, Pengfei and Zhou, Yuanchun and Fu, Yanjie},
  booktitle={Proceedings of the Thirty-Second International Joint Conference on Artificial Intelligence},
  pages={2086--2094},
  year={2023}
}

@inproceedings{sohl2015deep,
  title={Deep unsupervised learning using nonequilibrium thermodynamics},
  author={Sohl-Dickstein, Jascha and Weiss, Eric and Maheswaranathan, Niru and Ganguli, Surya},
  booktitle={International conference on machine learning},
  pages={2256--2265},
  year={2015},
  organization={pmlr}
}

@article{dhariwal2021diffusion,
  title={Diffusion models beat gans on image synthesis},
  author={Dhariwal, Prafulla and Nichol, Alexander},
  journal={Advances in neural information processing systems},
  volume={34},
  pages={8780--8794},
  year={2021}
}

@inproceedings{nichol2022glide,
  title={GLIDE: Towards Photorealistic Image Generation and Editing with Text-Guided Diffusion Models},
  author={Nichol, Alexander Quinn and Dhariwal, Prafulla and Ramesh, Aditya and Shyam, Pranav and Mishkin, Pamela and Mcgrew, Bob and Sutskever, Ilya and Chen, Mark},
  booktitle={International Conference on Machine Learning},
  pages={16784--16804},
  year={2022},
  organization={PMLR}
}

@inproceedings{kongdiffwave,
  title={DiffWave: A Versatile Diffusion Model for Audio Synthesis},
  author={Kong, Zhifeng and Ping, Wei and Huang, Jiaji and Zhao, Kexin and Catanzaro, Bryan},
  booktitle={International Conference on Learning Representations},
  year={2020}
}

@inproceedings{liu2023audioldm,
  title={AudioLDM: Text-to-Audio Generation with Latent Diffusion Models},
  author={Liu, Haohe and Chen, Zehua and Yuan, Yi and Mei, Xinhao and Liu, Xubo and Mandic, Danilo and Wang, Wenwu and Plumbley, Mark D},
  booktitle={International Conference on Machine Learning},
  pages={21450--21474},
  year={2023},
  organization={PMLR}
}

@article{li2022diffusion,
  title={Diffusion-lm improves controllable text generation},
  author={Li, Xiang and Thickstun, John and Gulrajani, Ishaan and Liang, Percy S and Hashimoto, Tatsunori B},
  journal={Advances in neural information processing systems},
  volume={35},
  pages={4328--4343},
  year={2022}
}

@inproceedings{gongdiffuseq,
  title={DiffuSeq: Sequence to Sequence Text Generation with Diffusion Models},
  author={Gong, Shansan and Li, Mukai and Feng, Jiangtao and Wu, Zhiyong and Kong, Lingpeng},
  booktitle={The Eleventh International Conference on Learning Representations},
  year={2022}
}

@inproceedings{gong2023diffuseq,
  title={DiffuSeq-v2: Bridging Discrete and Continuous Text Spaces for Accelerated Seq2Seq Diffusion Models},
  author={Gong, Shansan and Li, Mukai and Feng, Jiangtao and Wu, Zhiyong and Kong, Lingpeng},
  booktitle={Findings of the Association for Computational Linguistics: EMNLP 2023},
  pages={9868--9875},
  year={2023}
}

@article{yang2023generate,
  title={Generate what you prefer: Reshaping sequential recommendation via guided diffusion},
  author={Yang, Zhengyi and Wu, Jiancan and Wang, Zhicai and Wang, Xiang and Yuan, Yancheng and He, Xiangnan},
  journal={Advances in Neural Information Processing Systems},
  volume={36},
  pages={24247--24261},
  year={2023}
}

@inproceedings{shen2023diffusionner,
  title={DiffusionNER: Boundary Diffusion for Named Entity Recognition},
  author={Shen, Yongliang and Song, Kaitao and Tan, Xu and Li, Dongsheng and Lu, Weiming and Zhuang, Yueting},
  booktitle={Proceedings of the 61st Annual Meeting of the Association for Computational Linguistics (Volume 1: Long Papers)},
  pages={3875--3890},
  year={2023}
}

@inproceedings{wang2023diffusion,
  title={Diffusion recommender model},
  author={Wang, Wenjie and Xu, Yiyan and Feng, Fuli and Lin, Xinyu and He, Xiangnan and Chua, Tat-Seng},
  booktitle={Proceedings of the 46th International ACM SIGIR Conference on Research and Development in Information Retrieval},
  pages={832--841},
  year={2023}
}

@article{kazemi2018simple,
  title={Simple embedding for link prediction in knowledge graphs},
  author={Kazemi, Seyed Mehran and Poole, David},
  journal={Advances in neural information processing systems},
  volume={31},
  year={2018}
}

@inproceedings{liao2024gentkg,
  title={GenTKG: Generative Forecasting on Temporal Knowledge Graph with Large Language Models},
  author={Liao, Ruotong and Jia, Xu and Li, Yangzhe and Ma, Yunpu and Tresp, Volker},
  booktitle={Findings of the Association for Computational Linguistics: NAACL 2024},
  pages={4303--4317},
  year={2024}
}

@inproceedings{lee2023temporal,
  title={Temporal Knowledge Graph Forecasting Without Knowledge Using In-Context Learning},
  author={Lee, Dong-Ho and Ahrabian, Kian and Jin, Woojeong and Morstatter, Fred and Pujara, Jay},
  booktitle={Proceedings of the 2023 Conference on Empirical Methods in Natural Language Processing},
  pages={544--557},
  year={2023}
}

@article{wang2024large,
  title={Large language models-guided dynamic adaptation for temporal knowledge graph reasoning},
  author={Wang, Jiapu and Kai, Sun and Luo, Linhao and Wei, Wei and Hu, Yongli and Liew, Alan Wee-Chung and Pan, Shirui and Yin, Baocai},
  journal={Advances in Neural Information Processing Systems},
  volume={37},
  pages={8384--8410},
  year={2024}
}

@article{chiang2023vicuna,
  title={Vicuna: An open-source chatbot impressing gpt-4 with 90\%* chatgpt quality},
  author={Chiang, Wei-Lin and Li, Zhuohan and Lin, Ziqing and Sheng, Ying and Wu, Zhanghao and Zhang, Hao and Zheng, Lianmin and Zhuang, Siyuan and Zhuang, Yonghao and Gonzalez, Joseph E and others},
  journal={See https://vicuna. lmsys. org (accessed 14 April 2023)},
  volume={2},
  number={3},
  pages={6},
  year={2023}
}

@inproceedings{black2022gpt,
  title={GPT-NeoX-20B: An Open-Source Autoregressive Language Model},
  author={Black, Sidney and Biderman, Stella and Hallahan, Eric and Anthony, Quentin and Gao, Leo and Golding, Laurence and He, Horace and Leahy, Connor and McDonell, Kyle and Phang, Jason and others},
  booktitle={Proceedings of BigScience Episode\# 5--Workshop on Challenges \& Perspectives in Creating Large Language Models},
  pages={95--136},
  year={2022}
}

@article{touvronllama,
  title={LLaMA: Open and Efficient Foundation Language Models},
  author={Touvron, Hugo and Lavril, Thibaut and Izacard, Gautier and Martinet, Xavier and Lachaux, Marie-Anne and Lacroix, Timothee and Rozi{\`e}re, Baptiste and Goyal, Naman and Hambro, Eric and Azhar, Faisal and others},
  year={2023}
}

@article{luo2024chain,
  title={Chain of History: Learning and Forecasting with LLMs for Temporal Knowledge Graph Completion},
  author={Luo, Ruilin and Gu, Tianle and Li, Haoling and Li, Junzhe and Lin, Zicheng and Li, Jiayi and Yang, Yujiu},
  journal={CoRR},
  year={2024}
}

@inproceedings{pang2025improving,
  title={Improving Temporal Knowledge Graph Reasoning with Hierarchical Semantic-Aware Contrastive Learning},
  author={Pang, Renning and Liu, Yao and Gan, Yanglei and Dai, Tingting and Wang, Yashen and Shi, Xiaojun and Lan, Tian and Liu, Qiao},
  booktitle={Joint European Conference on Machine Learning and Knowledge Discovery in Databases},
  pages={376--394},
  year={2025},
  organization={Springer}
}

\newpage

\appendix

\section{Additional Related Work}

\subsection{Diffusion Models on Discrete Data}
Diffusion models (DMs) \cite{sohl2015deep} have emerged as a unifying generative paradigm, delivering state-of-the-art sample quality in image generation \cite{dhariwal2021diffusion,nichol2022glide} and audio \cite{kongdiffwave,liu2023audioldm}. While the original formulations operate in continuous Euclidean space, several lines of work now transpose the framework to discrete symbol sequences. Diffusion-LM \cite{li2022diffusion} projects word tokens into continuous embeddings and performing score matching within that continuous latent space, enabling effective non-autoregressive and controlled text generation, whereas DiffuSeq \cite{gongdiffuseq,gong2023diffuseq} adopts a discrete corruption process aligned with time indices and employs a transformer-based denoiser, enabling the generation of coherent sequences with arbitrary length. Beyond sequence generation, diffusion has also been adapted for structured prediction. DiffusionNER \cite{shen2023diffusionner}, for instance, formulates named-entity recognition as a boundary denoising task, gradually refining noisy span boundaries into consistent entity predictions. 

More recently, diffusion-based frameworks have been introduced into domain-specific applications that involve symbolic or sequential interactions. In recommendation systems, DiffRec \cite{wang2023diffusion} and DreamRec \cite{yang2023generate} leverage diffusion to model user–item interaction sequences. By injecting noise into observed behaviors and learning to reverse this process, these models uncover preference distributions and capture fine-grained uncertainty in recommendation outcomes. Such extensions highlight the versatility of diffusion paradigms as they are not only effective in dense continuous domains but also adaptable to discrete symbolic reasoning tasks, where noise-driven generative processes can reveal hidden structure and probabilistic dependencies.

\section{Preliminary}

\noindent\textbf{Definition 3. Diffusion Models for Discrete Data. }
Diffusion models (DMs) are probabilistic generative frameworks comprising two coupled Markov chains: a \textbf{forward diffusion} that gradually corrupts data with noise, and a \textbf{reverse denoising} that learns to recover the original samples. Although originally formulated for continuous domains, recent works have extended DMs to discrete sequences. We summarize the key components of discrete‐data diffusion below.

\textbf{Forward diffusion.} Given a discrete sequence ${w}$, we first embed it into continuous space via,

\begin{equation}
\footnotesize
\mathbf{x}_0 \sim \boldsymbol{q}\left(\mathbf{x}_0 \mid {w}\right)=\mathcal{N}\left(\mathbf{x}_0 ; \operatorname{Emb}({w}), \beta_0 \mathbf{I}\right),
\end{equation}
where $\operatorname{Emb}(\cdot) \in \mathbb{R}^d$ is an mapping function that projects each word into vector. $\beta_0>0$ controls the initial noise level. Thereafter, for $t \in \{0,1,\dots,T\}$, the latent states evolve according to :
\begin{equation} 
\footnotesize
\begin{aligned} 
q(\mathbf{x}t \mid \mathbf{x}{t-1}) &=\mathcal{N}\bigl(\mathbf{x}t;,\sqrt{\bar\alpha_t},\mathbf{x}{t-1},,(1-\bar\alpha_t)\mathbf{I}\bigr)\\
&=\sqrt{\bar\alpha_t},\mathbf{x}_{t-1};+;\sqrt{1-\bar\alpha_t},\boldsymbol\epsilon, \quad \boldsymbol\epsilon\sim\mathcal{N}(\mathbf{0},\mathbf{I}), \end{aligned} \label{eq:forward_diffusion} 
\end{equation}
where $\epsilon \sim \mathcal{N}(0,1)$ is a random Gaussian noise, and $\bar{\alpha}_t = \prod _{t'=1}^t, \alpha_{t'}\in (0,1)$ governs the noise schedule.

\textbf{Reverse denoising.} Starting from $\mathbf{x}_T \sim \mathcal{N}(\mathbf{0}, \mathbf{I})$, the reverse chain is parameterized by a neural network $f_\theta$ that predicts the Gaussian transition, 

\begin{equation}
\footnotesize
\begin{aligned}
p_\theta(\bold{x}_{t-1}|\bold{x}_t)&=\mathcal{N}(\bold{x}_{t-1};\mu_\theta(\bold{x}_t,t);\Sigma_\theta(\bold{x}_t,t)),
\end{aligned}
\end{equation}

where $\mu_\theta$ and $\Sigma_\theta$ are learnable functions of $(x_t, t)$. By iteratively sampling from $p_\theta$, one can recover an estimate $\widehat{\mathbf{x}}_0$. 

\textbf{Discrete rounding and training objective.} To map $\widehat{\mathbf{x}}_0$ back to tokens, we apply a soft-max based decoder, 

\begin{equation}
p_\theta\left(\boldsymbol{w} \mid \widehat{\mathbf{x}}_0\right)=\operatorname{Softmax}\left(\widehat{\mathbf{x}}_0\right).
\end{equation}

The end‐to‐end loss combines a continuous reconstruction term and a discrete rounding term,

\begin{equation}
\footnotesize
\begin{gathered}
\mathcal{L}_{ \text{simple}}(\mathbf{w})=\underset{q_\phi\left(\mathbf{x}_{0: T} \mid \mathbf{w}\right)}{\mathbb{E}}\left[\sum_{t=2}^T\left[\left\|\mathbf{x}_0-f_\theta\left(\mathbf{x}_t, t\right)\right\|^2\right]\right]+\\
\underset{q_\phi\left(\mathbf{x}_{0: 1} \mid \mathbf{w}\right)}{\mathbb{E}}\left[\left\|\operatorname{Emb}(\mathbf{w})-f_\theta\left(\mathbf{x}_1, 1\right)\right\|^2-\log p_\theta\left(\mathbf{w} \mid \mathbf{x}_0\right)\right] .
\label{eq:3}
\end{gathered}
\end{equation}

Where the first term trains the model to denoise from $t = 2$ to $T$, reducing continuous approximation errors, while the second term aligns the $t = 1$ denoised output with the original embedding and encourages accurate rounding to the discrete sequence.

\section{Experimental Setup}

\subsection{Dataset Statistics} \label{data stat}
To ensure consistency and comparability, we adhere to the chronological splitting protocol (80\% training, 10\% validation, and 10\% testing) established in prior studies by \cite{Li21Temporal,cai-etal-2024-predicting}.

\begin{table}[t]
\setlength{\tabcolsep}{1.5pt} % 减小列间距
\fontsize{8pt}{12pt}\selectfont % 设置特定字号和行距
\caption{The statistics of the datasets. $|E|$ and $|R|$ denote the number of unique entities and event types, respectively.}
\begin{tabular}{@{}ccccccc@{}}
\toprule
Datasets   & $|E|$        & $|R|$    & \textit{Train}  & \textit{Valid} & \textit{Test}        & \textit{Unseen Ratio} \\ \midrule
ICEWS14    & 6,869  & 230 & 74,845    & 8,514   & 7,371   & 58.43\%      \\
ICEWS18    & 23,033 & 256 & 373,018   & 45,995  & 49,545  & 55.69\%      \\
ICEWS05-15 & 10,094 & 251 & 368,868   & 46,302  & 46,159  & 39.82\%      \\
GDELT      & 7,691  & 240 & 1,734,399 & 238,765 & 305,241 & 43.72\%       \\ \bottomrule
\end{tabular}\label{stat}
\end{table}

\subsection{Baselines} \label{baseline}

\paragraph{Static Baselines:}
\begin{itemize}[leftmargin=*]
    \item DistMult \cite{yang2015embedding}, employs a bilinear scoring function, modeling triple plausibility via a diagonal relation matrix that captures pairwise interactions between subject and object embeddings.
    \item ConvE \cite{dettmers2018convolutional}, applies 2D convolution over reshaped entity and relation embeddings, followed by a projection layer to learn richer feature interactions for link prediction.
    \item RotatE \cite{sun2018rotate}, represents relations as complex-valued rotations in the embedding space, enabling the model to naturally encode and infer diverse relational patterns such as symmetry and inversion.
\end{itemize}

\paragraph{Interpolation Baselines:}

\begin{itemize}[leftmargin=*]
    \item TTransE \cite{leblay2018deriving}, explicitly models temporal dynamics by embedding entities and relations within a continuous time framework, using translation operations along the time dimension to capture their evolution.
    \item TA-DistMult \cite{garcia2018learning}, extends the DistMult scoring function with time‐aware embeddings, allowing the model to adapt relation parameters according to temporal context.
    \item DE-SimplE \cite{goel2020diachronic}, employs diachronic embeddings that parameterize entity and relation representations as functions of time, hence modeling progressive change across different timestamps.
\end{itemize}

\begin{figure*}[t]
	\centering
	\subfloat[Impact of $\tau$ on ICEWS14]{
		\includegraphics[scale=0.22]{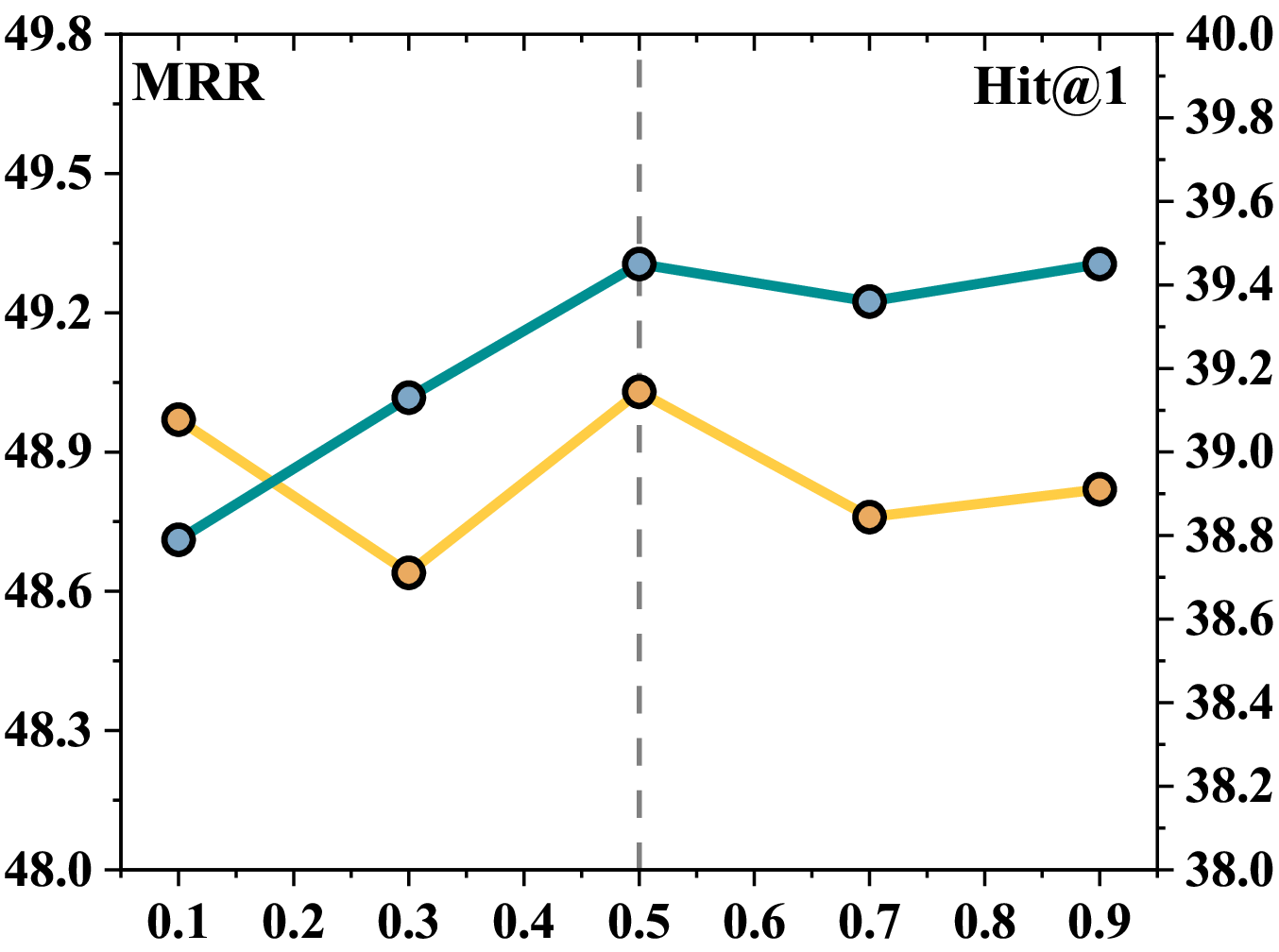}
	}%
	\subfloat[Impact of seq length on ICEWS14]{
		\includegraphics[scale=0.22]{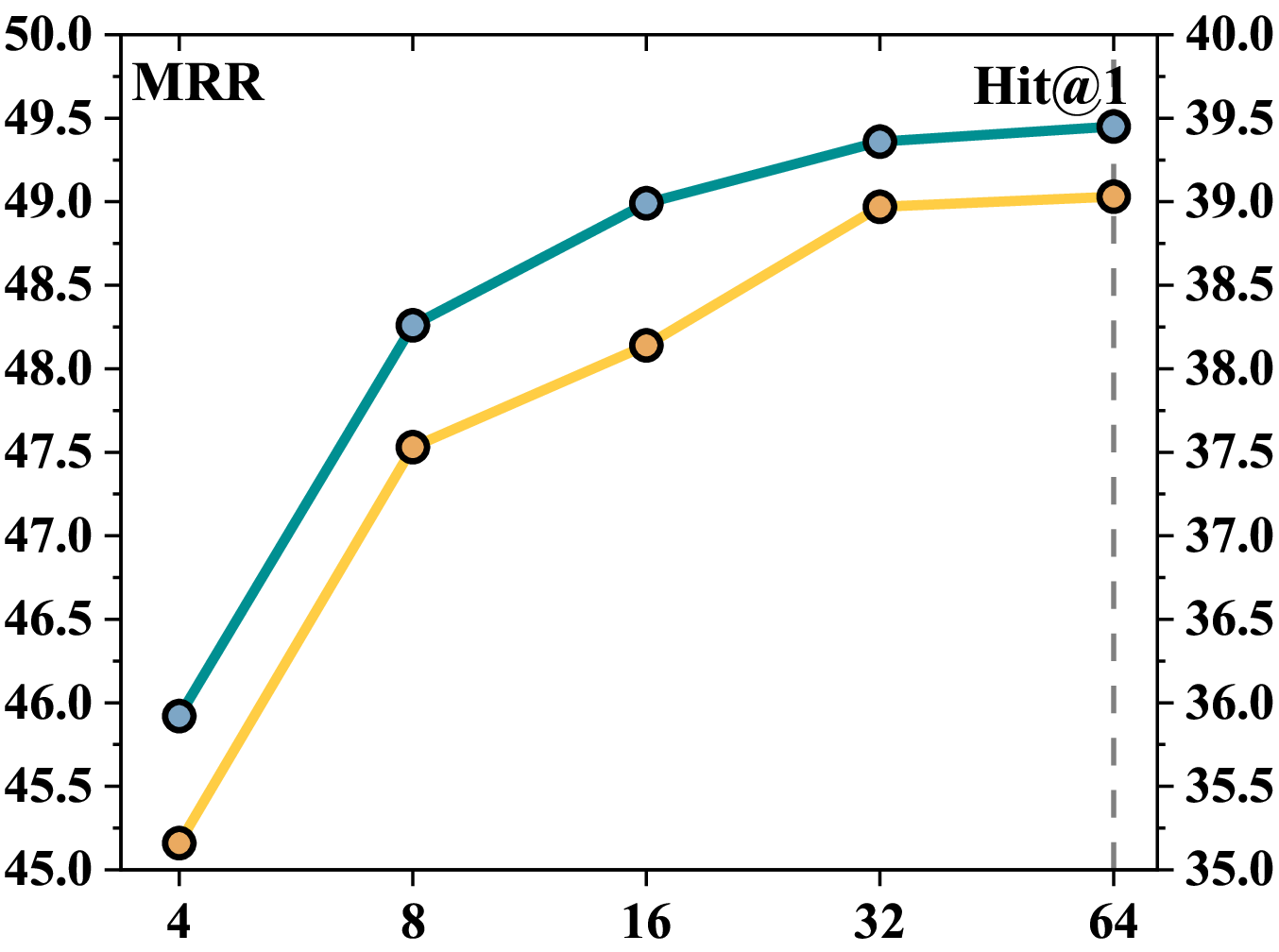}
	}%
 	\subfloat[Impact of noise scale on ICEWS14]{
		\includegraphics[scale=0.22]{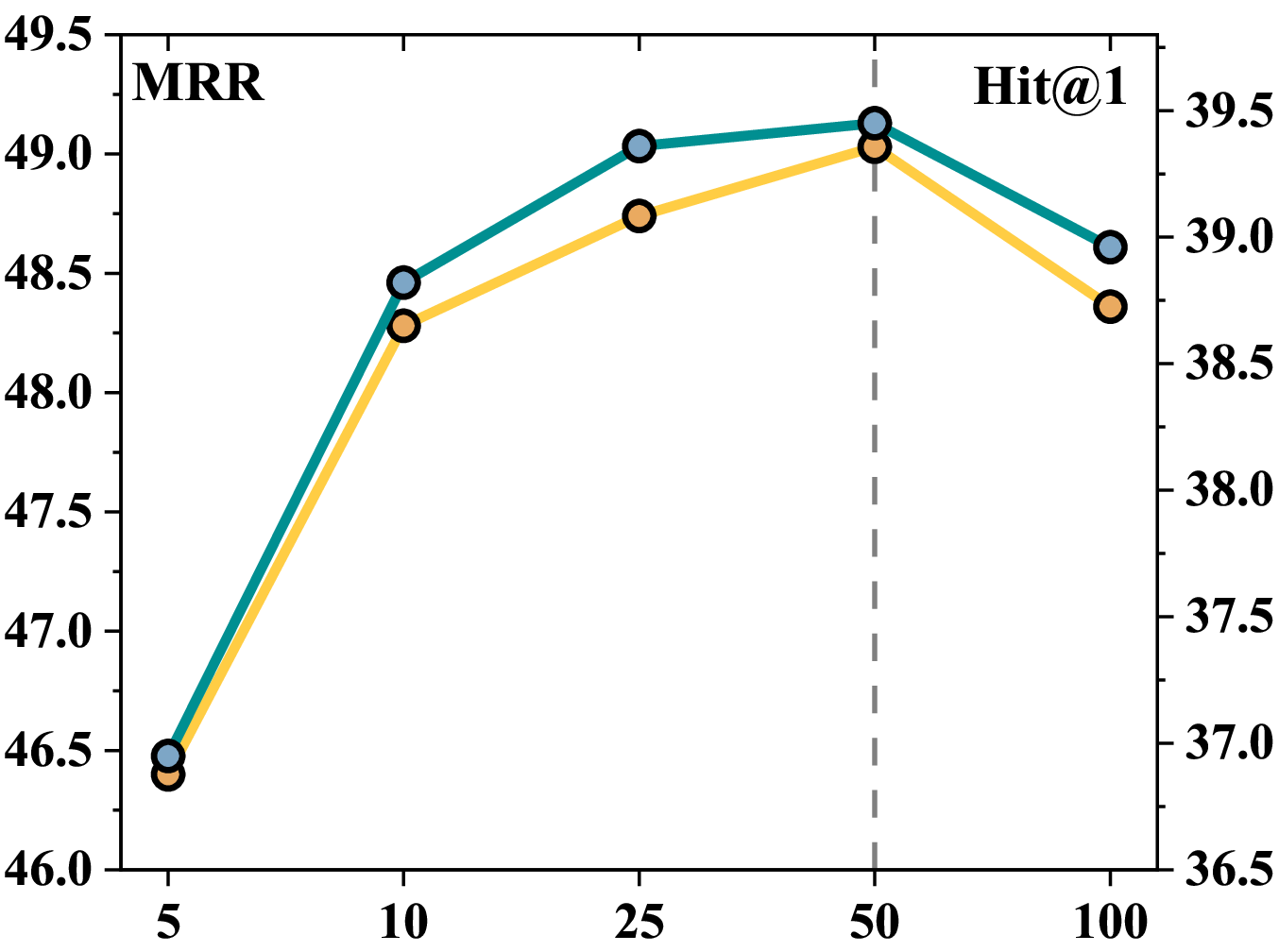}
	}%
 
 	\subfloat[Impact of $\tau$ on ICEWS18]{		
		\includegraphics[scale=0.22]{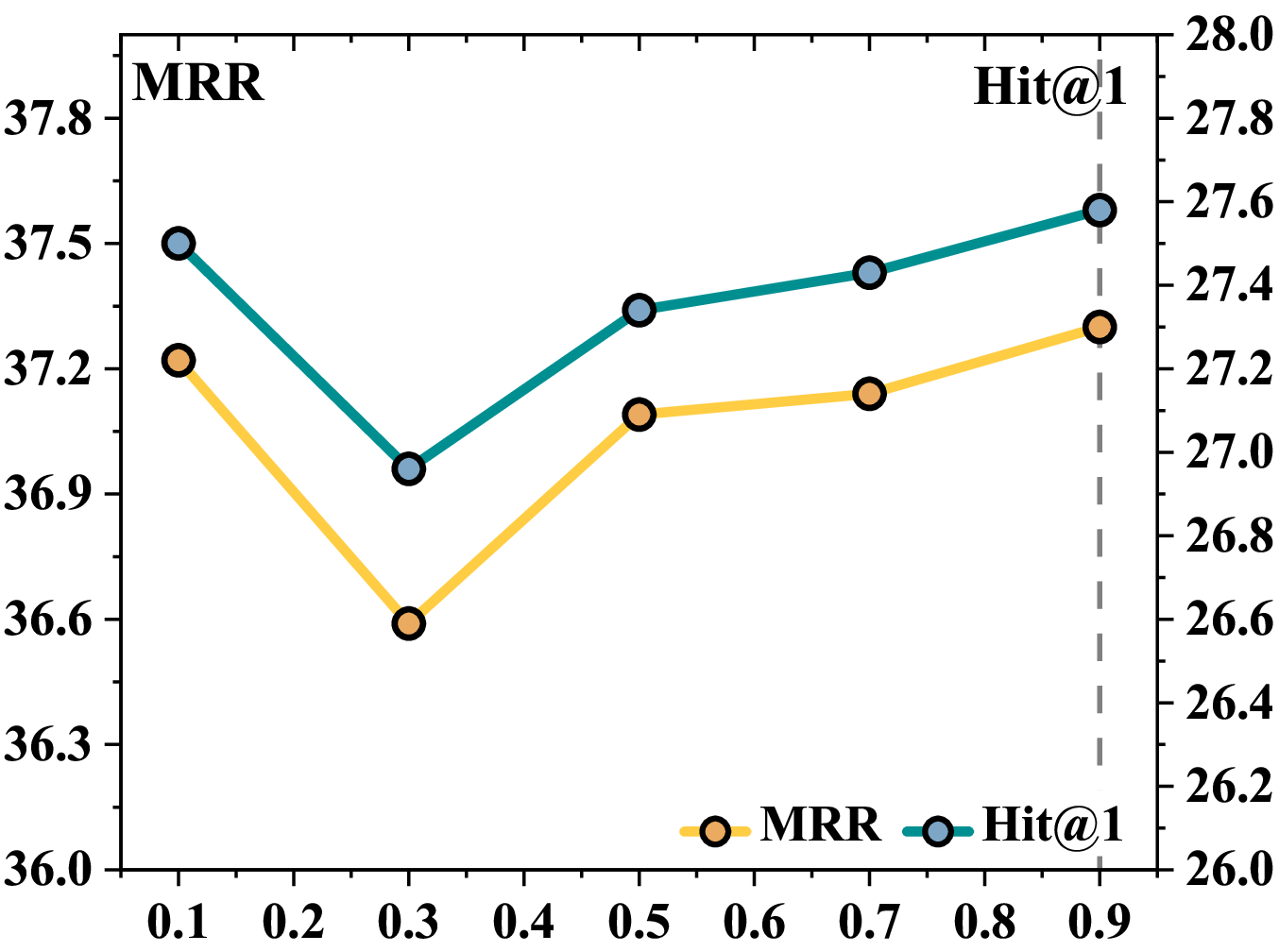}
	}%
	\subfloat[Impact of seq length on ICEWS18]{		
		\includegraphics[scale=0.22]{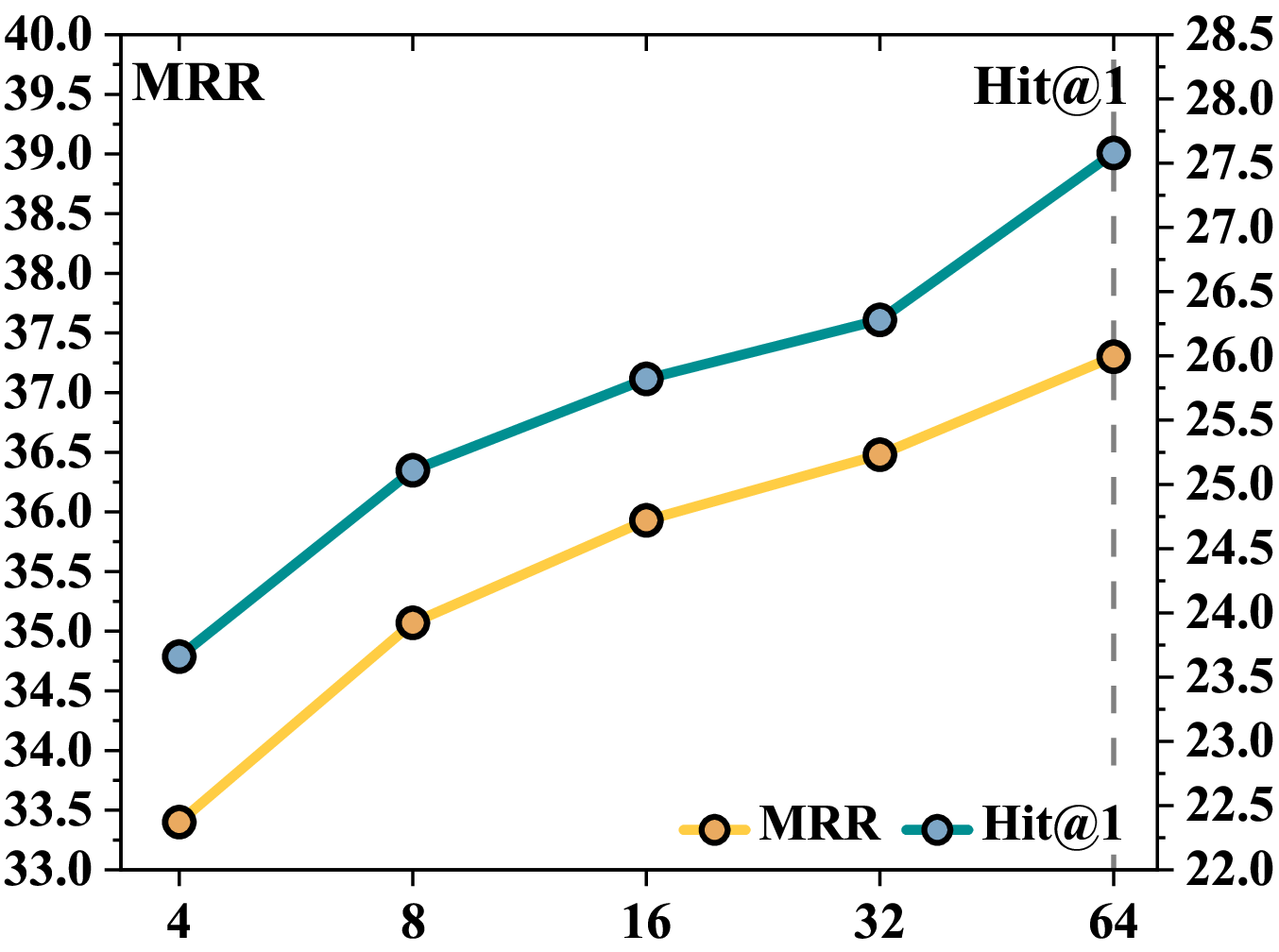}
	}%
	\subfloat[Impact of noise scale on ICEWS18]{		
		\includegraphics[scale=0.22]{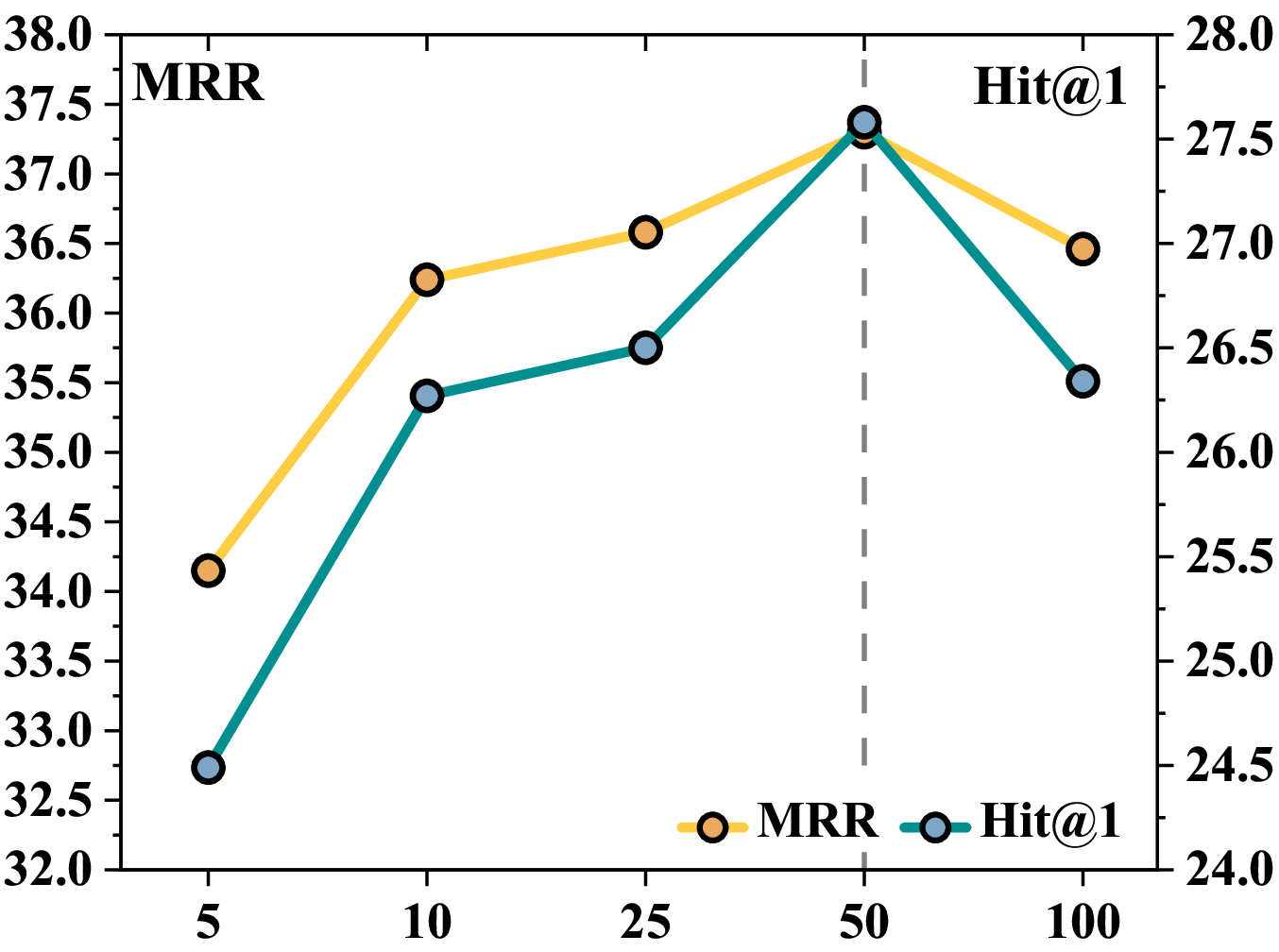}
	}%
	% \centering
	\caption{Sensitivity analysis on ICEWS14 and ICEWS18. }
	\label{fig5.6}
\end{figure*}

\paragraph{Extrapolation Baselines:}

\begin{itemize}[leftmargin=*]
    \item RE-NET \cite{jin2020recurrent}, integrates recurrent neural architectures with graph convolution to capture the sequential evolution of entities and predict future links.
    \item Re-GCN \cite{Li21Temporal}, employs a Recurrent Evolutionary GCN that recurrently updates entity and relation embeddings at each timestamp by propagating temporal signals through the KG.
    \item CyGNet \cite{zhu2021learning}, models cyclical temporal patterns, enabling the framework to learn periodic behaviors and extrapolate yet-unseen links.
    \item TITer \cite{sun2021timetraveler}, applies hierarchical transformations to entity embeddings, iteratively tracking their evolution to anticipate future TKG states.
    \item CEN \cite{li2022complex}, uses length-aware convolutional filters to extract multi-scale evolutionary patterns, with an online training strategy to handle temporal variability.
    \item TiRGN \cite{li2022tirgn}, leverages recurrent graph networks to encode dynamic relational structures, improving inference of unseen facts.
    \item HisMatch \cite{li2022hismatch}, aligns historical entity representations via a matching mechanism, enhancing prediction accuracy for future interactions.
    \item RETIA \cite{liu2023retia}, constructs a twin hyper-relation subgraph and evolutionarily aggregates adjacent entity and relation features for enriched message passing.
    \item CENET \cite{xu2023temporal}, incorporates contrastive learning objectives to strengthen dynamic representation learning within temporal graphs.
    \item CRAFT \cite{zhang2024modeling}, utilizes hypergraph convolutional aggregators to fuse coarse- and fine-grained historical contexts, yielding robust extrapolation.
    \item THCN \cite{chen2024thcn}, introduces temporal causal convolutional networks grounded in Hawkes processes to distinguish the relative importance of concurrent facts.
    \item DiffuTKG \cite{cai-etal-2024-predicting}, reframes TKG reasoning as a denoising diffusion process over entity embeddings to generate plausible future links.
    \item LogiQ \cite{chen2025enhancing}, augments diffusion-based generation with logical constraints, improving both accuracy and interpretability.
    \item CognTKE \cite{chen2025cogntke}, integrates cognitively inspired symbolic priors into embedding learning, enabling more transparent and reliable extrapolation.
\end{itemize}

\section{Experimental Analysis}

\subsection{Sensitivity Analysis}
In this section, we evaluate NADEx's sensitivity to three key hyperparameters on the ICEWS14 and ICEWS18 benchmarks in terms of MRR and Hit@1, including the impact of temperature coefficient $\tau$, different lengths of event sequence and different scale of noise in the forward process. Our findings are summarized in Figure \ref{fig5.6}. 

\noindent\textbf{Impact of Temperature Coefficient. }
The temperature coefficient $\tau$ in our log‑likelihood ranking loss controls the sharpness of the softmax distribution over candidate entities. From Figure \ref{fig5.6}(a) on ICEWS14, we observe that both MRR and Hit@1 improve as $\tau$ increases from 0.1 to 0.5. Beyond this point, performance plateaus or slightly dips. In contrast, on the more temporally complex ICEWS18 (Figure \ref{fig5.6}(d)), performance initially drops at $\tau$ = 0.3, but then steadily rises across the range $\tau$ = 0.5 to $\tau$ = 0.9, with optimal MRR and Hit@1 achieved at $\tau$ = 0.9.

\noindent\textbf{Impact of Event Sequence Length. }
The length of the historical event sequence notably influences NADEx's predictive power, as shown in Figures \ref{fig5.6}(b) and (e). On ICEWS14, increasing the input window from 4 to 8 events yields a substantial jump, indicating that the model benefits from additional context. Further extending the sequence to 16, 32, and 64 events provides diminishing but still positive returns. The plateau around 32 to 64 suggests extra history adds limited new information on ICEWS14. ICEWS18 exhibits a similar pattern, with steady MRR and Hit@1 increases up to 64 events, and a relative acceleration in improvement beyond 32, reflecting its more complex temporal dynamics. It is worth-noting that we restrict our analysis to a maximum sequence length of 64 to balance predictive gains against the computational overhead on a single NVIDIA A100 (80 GB) GPU.

\noindent\textbf{Impact of Noise Scale. }
We further investigate the influence of the maximum noise scale in the forward diffusion process on NADEx's capacity to model and denoise intricate temporal dynamics. Figures \ref{fig5.6}(c) and (f) report MRR and Hit@1 as the perturbation magnitude applied to the target embedding is varied. When the scale is small (5–10), the injected noise is insufficient to exploit the generative power of diffusion, resulting in sub-optimal accuracy. Performance improves markedly at a moderate scale of 25 and reaches its optimum at 50 on both ICEWS14 and ICEWS18, indicating that a balanced level of stochasticity best enriches the denoising signal. Conversely, very large perturbations drive the latent representation far from the data manifold; the reverse process must then traverse a longer, noisier trajectory, weakening gradient signals and making accurate reconstruction substantially more difficult.

\end{document}